\documentclass{article}

\PassOptionsToPackage{numbers, compress}{natbib}

\usepackage[preprint]{neurips_2026}

\usepackage[utf8]{inputenc} 
\usepackage[T1]{fontenc}    
\usepackage{hyperref}       
\usepackage{url}            
\usepackage{booktabs}       
\usepackage{amsfonts}       
\usepackage{amsmath}
\usepackage{nicefrac}       
\usepackage{microtype}      
\usepackage{xcolor}         
\usepackage{graphicx}
\usepackage{multirow}
\usepackage{enumitem}
\usepackage{float}          
\usepackage{placeins}       
\usepackage{tikz}
\usetikzlibrary{arrows.meta,positioning,fit,calc,backgrounds}
\usepackage[capitalise]{cleveref} 

\definecolor{pretrainedfill}{HTML}{E5EDF5}
\definecolor{pretraineddraw}{HTML}{2B5B84}
\definecolor{originalfill}{HTML}{E2F0EA}
\definecolor{originaldraw}{HTML}{27705B}

\definecolor{secondbest}{HTML}{2E7D32}
\newcommand{\ssb}[1]{\textit{\textcolor{secondbest}{#1}}}

\title{High-Throughput Generative Video Super-Resolution on a Single Consumer GPU}

\author{%
  Georgios Panaretou \\
  Ethical AI Novelties \\
  Limassol, Cyprus \\
  \texttt{ggpanaretou@gmail.com} \\
  \And
  Sotirios Chatzis \\
  Cyprus University of Technology \\
  Limassol, Cyprus \\
  \texttt{sotirios.chatzis@cut.ac.cy} \\
}

\begin{document}

\maketitle

\begin{abstract}
Generative video super-resolution (VSR) can recover visually compelling high-frequency detail by leveraging diffusion priors, but iterative sampling and heavy spatio-temporal coupling make practical deployments difficult. We study a different operating point: practical HD generative VSR on a single high-end consumer GPU without end-to-end retraining. To this end, we present a zero-shot framework built entirely from pretrained components, where ``zero-shot'' denotes inference-time composition without benchmark-specific finetuning.

The proposed system decouples spatial enhancement from temporal stabilization. A time-distilled one-step diffusion backbone handles spatial upscaling, while temporal coherence is promoted by a lightweight pretrained flow-based interpolator that synthesizes intermediate frames between super-resolved keyframes. 
To further improve scalability, we introduce a resolution-aware hybrid attention scheme that, in a zero-shot manner, replaces selected high-resolution global-attention layers with neighborhood attention while preserving global attention in semantically critical low-resolution layers, while also improving temporal consistency.

Experiments show that the proposed framework reaches up to 32.66 FPS at 1280$\times$720 resolution under interleaved inference on a single NVIDIA RTX 4090. 
Strong diffusion-based comparative baselines can provide higher quality at much higher computational and memory costs, resulting in much lower throughput. 
At 3840$\times$2160, the proposed attention substitution can yield up to a 44.7\% speedup over the original attention implementation. 
Across standard VSR benchmarks, the method consistently shifts the speed-quality trade-off decisively toward runtime efficiency, trading reconstruction fidelity for strong no-reference perceptual quality, while remaining reasonably competitive against generative approaches in temporal consistency. It also exposes clear limitations in handling more challenging motion, which a simple motion gate partially mitigates. Code will be released upon acceptance.
\end{abstract}

\section{Introduction}
\label{sec:intro}

\begin{figure}[t]
  \centering
  \includegraphics[width=\linewidth]{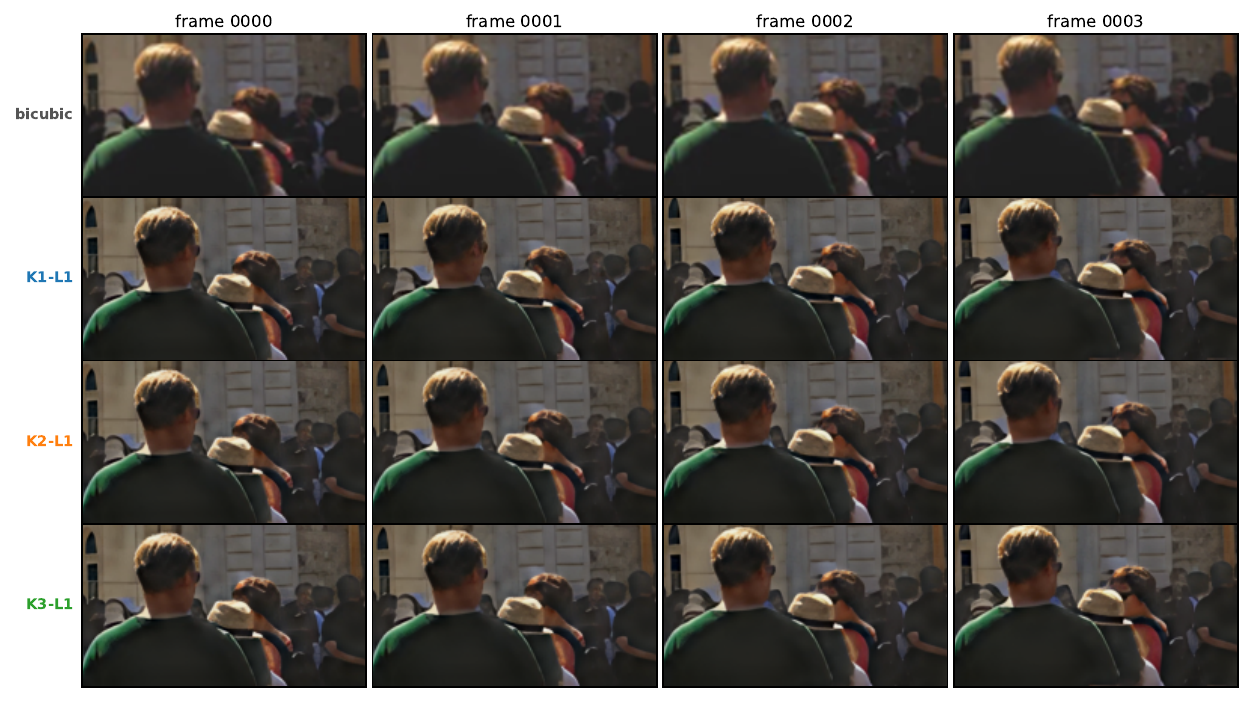}
  \caption{Our Video Super-Resolution compared to bicubic interpolation on consecutive frames. $K$ denotes the super-resolution frequency; in-between frames are interpolated. $L$ denotes the attention blocks replaced with NATTEN.}
  \label{fig:quantitative_image_comparison}
\end{figure}

Video super-resolution (VSR) has progressed from distortion-oriented convolutional and recurrent architectures toward perceptually driven generative models. 
Classical systems based on deformable alignment, recurrent propagation, or reconstruction-heavy objectives can achieve strong PSNR and SSIM. Still, they often recover overly smooth textures and under-represent fine visual detail \cite{Ledig_2017_CVPR,wang2019edvr,chan2022basicvsrplusplus}. 
By contrast, diffusion models can synthesize sharper and more realistic high-frequency structure by exploiting strong image priors \cite{Ho_NEURIPS2020,Rombach_2022_CVPR}. 
Recent diffusion-based VSR systems confirm this promise, but they typically inherit the cost of multi-step denoising together with additional temporal modules, which makes real-time operation difficult \cite{zhou2024upscaleavideo,yeh2024diffir2vr,sun2025onestepdiffusiondetailrichtemporally}.

The main bottleneck is not only iterative sampling itself, but also the way spatial generation and temporal consistency are usually coupled. 
Strong perceptual quality often comes with spatio-temporal attention, recurrent latent propagation, or jointly trained video-specific refinement, all of which increase latency and memory pressure. 

In this paper, we explore a more compute-predictable alternative. 
Instead of fully generative temporal modeling, we decouple spatial enhancement from temporal stabilization. 
A pretrained one-step diffusion model produces perceptually rich super-resolved keyframes, while a lightweight pretrained interpolation network fills in intermediate frames. 
We further increase throughput and improve temporal consistency by replacing selected shallow global-attention blocks with neighborhood attention in a zero-shot manner.

The resulting system requires no end-to-end retraining, no benchmark-specific finetuning, and no video-specific optimization. 
Across standard benchmarks, it delivers a clear speed--perception trade-off: strong no-reference perceptual quality, improved temporal stability under interleaved inference, and substantially higher throughput than diffusion-based baselines under matched single-batch HD evaluation.
Our main contributions are as follows:
\begin{itemize}[nosep]
    \item \textbf{A decoupled generative VSR framework.} We propose a modular zero-shot pipeline that combines one-step diffusion-based spatial enhancement with pretrained flow-guided temporal interpolation, enabling practical HD VSR on a single consumer GPU.
    \item \textbf{A compute-aware, motion-adaptive temporal inference strategy.} We introduce interleaved keyframe super-resolution with intermediate frame synthesis, exposing explicit efficiency--quality trade-offs through the keyframe interval $K$ and reaching up to 32.66 FPS at 1280$\times$720 when using batching; we further introduce a lightweight, training-free motion gate that detects unreliable interpolation on a per-frame basis and automatically reroutes affected frames to the generative backbone instead of interpolating them, at the cost of throughput.
    \item \textbf{A zero-shot hybrid attention substitution.} Building on the locality principle established in HDiT, we retain global attention where global semantics matter most and replace selected high-resolution attention layers with neighborhood attention where the compute burden is largest, obtaining up to a 44.7\% speedup at 4K.
    \item \textbf{A careful characterization of the operating point.} Across four benchmarks, we show the attention substitution induces a clear perception--distortion--temporal trade-off, with temporal stability uniquely optimized at an intermediate layer count; we further show that the temporal degradation on challenging high-motion sequences is a localized, motion-triggered failure mode that our gating mechanism can automatically detect and partially mitigate.
\end{itemize}

\section{Related Work}

\textbf{Regression-based VSR.}
Before generative restoration became prominent, VSR was dominated by distortion-oriented methods trained with pixel-wise or reconstruction-heavy losses. EDVR \cite{wang2019edvr} aligns frames with deformable convolutions, while BasicVSR++ \cite{chan2022basicvsrplusplus} improves information propagation with bidirectional recurrent processing \cite{wang2019edvr,chan2022basicvsrplusplus}. These methods remain strong on PSNR and SSIM, but their outputs are often perceptually conservative, especially on difficult textures.

\textbf{Diffusion-based image and video restoration.}
Diffusion models and latent diffusion models provide powerful generative priors for restoration \cite{Ho_NEURIPS2020,Rombach_2022_CVPR}. Extending them to video requires additional temporal modeling. Upscale-A-Video introduces temporal attention and latent propagation for temporally coherent real-world VSR, while VideoLDM augments image backbones with temporal layers \cite{zhou2024upscaleavideo,Blattmann_2023_CVPR}. DiffIR2VR-Zero avoids retraining by combining image restoration diffusion priors with optical flow and token merging \cite{yeh2024diffir2vr}. These methods demonstrate strong perceptual capability, but their multi-step diffusion backbones and temporal coupling still impose substantial inference cost.

\textbf{One-step diffusion and efficient generative inference.}
A complementary direction compresses diffusion sampling into one or a few steps. 
Consistency-based methods and adversarial diffusion compression aim to preserve generative quality while drastically reducing inference latency \cite{pmlr-v202-song23a,chen2025adversarial}. 
For super-resolution, one-step image restoration models such as OSEDiff and AdcSR show that pretrained diffusion priors can be distilled into fast feed-forward architectures \cite{Ma2024OneStepDiffusionSR,chen2025adversarial}. 
Most recently, AdcVSR~\cite{adcvsr2026} extends the ADC compression philosophy underlying our own spatial backbone to the video setting, distilling a 3D spatio-temporal DiT teacher into a pruned 2D backbone augmented with temporal convolutions and a dual-head adversarial objective. 

\textbf{Efficient attention at high resolution.}
Even in one-step backbones, global self-attention becomes a major bottleneck as spatial resolution grows. FlashAttention reduces memory traffic and improves implementation efficiency, but it does not change the underlying quadratic dependence on sequence length \cite{dao2023flashattention2}. 
Neighborhood attention offers a hardware-friendly sparse alternative with localized receptive fields and linear complexity in the number of pixels for a fixed window size \cite{hassani2025generalized}. 
Closest to our attention design is the Hourglass Diffusion Transformer (HDiT)~\cite{crowson2024hourglasshdit}, which trains a diffusion transformer with neighborhood attention in high-resolution layers and global attention in low-resolution layers.

\textbf{Positioning of our design.}
The proposed framework sits at the intersection of these threads. 
It is not an end-to-end video diffusion model, nor merely a fast image super-resolver applied frame by frame. 
Instead, it combines a distilled generative image prior, a motion-based temporal module, and a targeted architectural acceleration mechanism into a single pretrained-only system. 

\begin{figure}[t]
    \centering
    \footnotesize
    \resizebox{0.985\linewidth}{!}{%
\begin{tikzpicture}[
    >=Latex,
    x=1cm, y=1cm,
    font=\small,
    line cap=round,
    line join=round,
    module/.style={draw=black!24, line width=0.95pt, rounded corners=4.8pt, align=center, minimum height=1.40cm, inner sep=5pt, fill=white},
    pretrained/.style={module, draw=pretraineddraw, fill=pretrainedfill},
    original/.style={module, draw=originaldraw, fill=originalfill},
    chip/.style={rounded corners=8pt, inner sep=3.5pt, font=\scriptsize\bfseries},
    tagpre/.style={rounded corners=7pt, inner sep=2.5pt, font=\scriptsize\bfseries, draw=pretraineddraw, fill=white, text=pretraineddraw},
    tagour/.style={rounded corners=7pt, inner sep=2.5pt, font=\scriptsize\bfseries, draw=originaldraw, fill=white, text=originaldraw},
    panel/.style={draw=black!15, line width=0.75pt, rounded corners=7pt, fill=black!2},
    paneltitle/.style={font=\bfseries\scriptsize, text=black!76},
    arr/.style={-{Latex[length=2.8mm,width=1.8mm]}, line width=1.05pt, draw=black!72},
    frame/.style={draw=black!36, rounded corners=1.4pt, minimum width=0.72cm, minimum height=0.62cm, fill=white, font=\scriptsize},
    srframe/.style={draw=originaldraw, fill=originalfill, rounded corners=1.4pt, minimum width=0.72cm, minimum height=0.62cm, font=\scriptsize},
    intframe/.style={draw=pretraineddraw, fill=pretrainedfill, rounded corners=1.4pt, minimum width=0.72cm, minimum height=0.62cm, font=\scriptsize},
    smalllab/.style={font=\scriptsize, text=black!60}
]

    \node[chip, draw=pretraineddraw, fill=pretrainedfill, anchor=west] (leg1) at (0.0, 5.0) {pretrained};
    \node[chip, draw=originaldraw, fill=originalfill, right=0.2cm of leg1] (leg2) {ours};

    \node[panel, minimum width=3.8cm, minimum height=1.8cm, anchor=west] (input) at (0.0, 1.3) {};
    \node[anchor=north, font=\bfseries\footnotesize] at ($(input.north)+(0,-0.15)$) {Low-resolution clip};
    \node[frame] (in0) at ($(input.center)+(-1.3,-0.1)$) {$x_0$};
    \node[frame, right=0.15cm of in0] (in1) {$x_1$};
    \node[frame, right=0.15cm of in1] (in2) {$x_2$};
    \node[frame, right=0.15cm of in2] (in3) {$x_3$};
    \node[anchor=south, smalllab] at ($(input.south)+(0,0.05)$) {$\{x_0,x_1,\ldots,x_T\}$};

    \node[original, minimum width=3.2cm, minimum height=1.4cm, anchor=west] (policy) at (4.6, 1.3)
        {\\[0.5ex]Interleaved keyframe policy\\[0.5ex]$K\in\{1,2,3\}$};
    \node[tagour, anchor=north west] at ($(policy.north west)+(0.08,-0.08)$) {ours};
    \node[chip, draw=originaldraw, fill=white, anchor=south] at ($(policy.north)+(0,0.15)$) {invoke generator every $K$-th frame};
    \node[smalllab, anchor=north] at ($(policy.south)+(0,-0.10)$) {$K=3$: SR $\rightarrow$ Int $\rightarrow$ Int $\rightarrow$ SR $\cdots$};

    \node[panel, minimum width=6.2cm, minimum height=2.2cm, anchor=west] (spbox) at (8.6, 3.5) {};
    \node[paneltitle, anchor=north west] at ($(spbox.north west)+(0.15,-0.15)$) {Sparse generative path (keyframes and motion-gated fallback)};
    \node[pretrained, minimum width=2.6cm, anchor=west] (adcsr) at ($(spbox.west)+(0.3,-0.1)$)
        {\\[0.5ex]AdcSR\\spatial generator};
    \node[tagpre, anchor=north west] at ($(adcsr.north west)+(0.08,-0.08)$) {pretrained};
    \node[original, minimum width=2.0cm, minimum height=1.4cm, right=0.6cm of adcsr] (hyb)
        {\\[0.5ex]Zero-shot hybrid\\attention};
    \node[tagour, anchor=north west] at ($(hyb.north west)+(0.08,-0.08)$) {ours};

    \node[panel, minimum width=6.2cm, minimum height=2.2cm, anchor=west] (tpbox) at (8.6, -0.9) {};
    \node[paneltitle, anchor=north west] at ($(tpbox.north west)+(0.15,-0.15)$) {Temporal fill-in path (between keyframes)};
    \node[pretrained, minimum width=2.6cm, anchor=west] (rife) at ($(tpbox.west)+(0.3,-0.1)$)
        {\\[0.5ex]RIFE\\interpolator};
    \node[tagpre, anchor=north west] at ($(rife.north west)+(0.08,-0.08)$) {pretrained};
    \node[original, minimum width=2.0cm, minimum height=1.4cm, right=0.6cm of rife] (interp)
        {\\[0.5ex]Arbitrary-timestep\\interpolation};
    \node[tagour, anchor=north west] at ($(interp.north west)+(0.08,-0.08)$) {ours};

    \node[original, minimum width=2.6cm, minimum height=1.4cm, anchor=west] (merge) at (15.6, 1.3)
        {\\[0.5ex]Temporal\\assembly};
    \node[tagour, anchor=north west] at ($(merge.north west)+(0.08,-0.08)$) {ours};

    \node[panel, minimum width=3.8cm, minimum height=1.8cm, anchor=west] (output) at (19.0, 1.3) {};
    \node[anchor=north, font=\bfseries\footnotesize] at ($(output.north)+(0,-0.15)$) {High-resolution clip};
    \node[srframe] (out0) at ($(output.center)+(-1.3,-0.1)$) {$y_0$};
    \node[intframe, right=0.15cm of out0] (out1) {$\hat y_1$};
    \node[intframe, right=0.15cm of out1] (out2) {$\hat y_2$};
    \node[srframe, right=0.15cm of out2] (out3) {$y_3$};
    \node[anchor=south, smalllab] at ($(output.south)+(0,0.05)$) {$\{y_0,y_1,\ldots,y_T\}$};

    \draw[arr] (input.east) -- (policy.west);

    \draw[arr] (policy.east) -- ++(0.4,0) |- (adcsr.west);

    \draw[arr] (adcsr.east) -- (hyb.west);
    \draw[arr] (rife.east) -- (interp.west);

    \draw[arr] (hyb.east) -- ++(1.1,0) |- (merge.north);
    \draw[arr] (interp.east) -- ++(1.1,0) |- (merge.south);

    \draw[arr] (merge.east) -- (output.west);

    \draw[arr] (hyb.south) -- ++(0, -1.2) coordinate (midpt)
        -- node[fill=white, font=\tiny, inner sep=1.2pt, text=black!65] {SR-keyframes} (rife.north |- midpt)
        -- (rife.north |- tpbox.north);

    
    \node[font=\tiny, fill=white, inner sep=1.2pt, text=black!65, anchor=south] at ($(adcsr.west) + (-0.5, 0.1)$) {keyframes};
    \node[font=\tiny, fill=white, inner sep=1.2pt, text=black!65, anchor=west] at ($(hyb.east) + (1.1, -0.6)$) {SR keyframes};
    \node[font=\tiny, fill=white, inner sep=1.2pt, text=black!65, anchor=west] at ($(interp.east) + (1.1, 0.6)$) {intermediate frames};

    \node[draw=originaldraw, dashed, line width=1.0pt, rounded corners=6pt, fit=(policy)(spbox)(tpbox)(merge), inner sep=0.4cm] (novel) {};
    \node[chip, draw=originaldraw, fill=white, anchor=south west] at ($(novel.north west)+(0.2,-0.01)$)
        {novelty = composition + scheduling + selective attention substitution};

\end{tikzpicture}%
    }
    \caption{Proposed Approach. The system reuses two pretrained assets, namely AdcSR for sparse keyframe super-resolution and RIFE for temporal interpolation. At the same time, the original contribution of this work lies in how these components are combined: a motion-aware interleaved keyframe policy controls when the generative backbone is invoked, zero-shot hybrid attention accelerates the high-resolution layers of the spatial generator, and temporal assembly merges super-resolved keyframes with interpolated intermediate frames.}
    \label{fig:pipeline}
\end{figure}
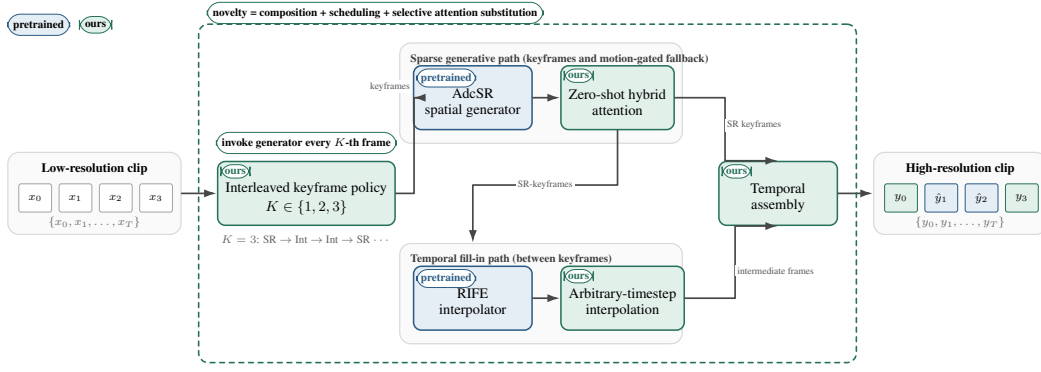

\section{Proposed Approach}
\subsection{System Overview}
Given a low-resolution video sequence $\mathcal{V}_{\mathrm{LR}}=\{x_0,x_1,\dots,x_T\}$, our goal is to generate a high-resolution sequence $\mathcal{V}_{\mathrm{HR}}=\{y_0,y_1,\dots,y_T\}$ with high perceptual fidelity and temporal coherence. 
The overall pipeline is summarized in Figure~\ref{fig:pipeline}. 
Our design contributes an interleaved, motion-gated keyframe policy that decouples sparse generative super-resolution from dense flow-based interpolation, and a resolution-aware hybrid attention scheme that transplants a locality principle previously validated only at training time into an already-distilled, pretrained one-step SR backbone, without any fine-tuning. 
We instantiate the spatial and temporal components with a one-step diffusion super-resolver $\mathcal{G}_{\mathrm{SR}}$ (AdcSR) and a real-time interpolation network $\mathcal{G}_{\mathrm{Int}}$ (RIFE); we use the term \emph{zero-shot} throughout to describe this inference-time composition, without end-to-end retraining, benchmark-specific finetuning, or additional joint optimization.

\subsubsection{Interleaved keyframe strategy}
We introduce a keyframe interval $K$, where frames whose indices satisfy $t \bmod K = 0$ are treated as keyframes and are processed by the generative backbone. 
Frames in between are synthesized by interpolation using the nearest super-resolved keyframes. 
Formally,
\begin{equation}
    y_t=
    \begin{cases}
        \mathcal{G}_{\mathrm{SR}}(x_t), & t \bmod K = 0,\\
        \mathcal{G}_{\mathrm{Int}}\big(y_{t_{\mathrm{prev}}},y_{t_{\mathrm{next}}},\tau_t\big), & \text{otherwise},
    \end{cases}
\end{equation}
where $t_{\mathrm{prev}}$ and $t_{\mathrm{next}}$ are the neighboring keyframe indices and
\begin{equation}
    \tau_t = \frac{t-t_{\mathrm{prev}}}{t_{\mathrm{next}}-t_{\mathrm{prev}}}
\end{equation}
indicates the relative interpolation timestamp.

We study three operating points:
\begin{itemize}
    \item \textbf{$K{=}1$:} every frame is super-resolved by $\mathcal{G}_{\mathrm{SR}}$;
    \item \textbf{$K=2$:} one frame is super-resolved, and one intermediate frame is interpolated;
    \item \textbf{$K=3$:} one frame is super-resolved, and two intermediate frames are interpolated.
\end{itemize}
This design keeps the system simple and compute-predictable: the expensive generator is invoked sparsely, while the cheaper interpolation branch dominates throughput for larger $K$.

\subsection{Generative Spatial Upscaling}
Our spatial generator is AdcSR \cite{chen2025adversarial}, a one-step diffusion-based super-resolution model distilled from Stable Diffusion priors. We select this backbone because it preserves much of the perceptual richness associated with diffusion models while avoiding iterative denoising during inference.

\textbf{Structural compression.}
AdcSR inherits the OSEDiff design philosophy \cite{Ma2024OneStepDiffusionSR} but removes components that are unnecessary for image-to-image super-resolution. In particular, the encoder-side VAE path and the text-conditioning branch are omitted, and the low-resolution input is mapped into the latent space via PixelUnshuffle rather than a heavy encoder. The U-Net is also channel-pruned, which materially reduces compute and memory cost.

\textbf{One-step inference.}
Because the model is distilled into a single forward pass, spatial super-resolution becomes a feed-forward operation rather than a many-step sampling procedure. This makes it viable as the keyframe generator in our pipeline.

\subsection{Resolution-Aware Hybrid Attention}
\label{sec:natten}
Even with one-step inference, attention remains a major computational bottleneck at high spatial resolution. For a feature map with $N=H\times W$ tokens and channel dimension $C$, standard self-attention has complexity
\begin{equation} \label{attention_equation}
    \Omega_{\mathrm{SA}} = 4NC^2 + 2N^2C
\end{equation}
So the quadratic term becomes dominant as resolution grows. This bottleneck is especially pronounced in the shallow high-resolution layers of the U-Net.

We therefore adopt a hybrid attention design. In low-resolution bottleneck layers, where semantic context is global, and the token count is small, we keep standard global attention. 
In shallow, high-resolution layers, where feature interactions are predominantly local, and token counts are large, we replace selected global attention layers with neighborhood attention \cite{hassani2025generalized}. 
For a fixed neighborhood window size $\omega$, the resulting complexity is
\begin{equation}
    \Omega_{\mathrm{NA}} = 4NC^2 + 2N\omega^2C,
\end{equation}
which is linear in $N$ for fixed $\omega$.

\textbf{Zero-shot substitution.}
The substitution is applied without finetuning. In practice, we reuse the pretrained projection layers and restrict their interactions to local windows.
An ablation study across UDM10, SPMCS, Vid4, and REDS4 is available in Appendix~\ref{natten_ablation_study}, in which we study how replacing layers in the encoder and decoder of the generative backbone's UNet affects the output.

\textbf{Implementation details.} 
We use a NATTEN window size of $\omega=7$ throughout all experiments.
AdcSR's U-Net has $5$ layers, including two encoder layers and three decoder layers, operating on high-resolution spatial feature maps.
Motivated by the design intuition in Section~\ref{sec:natten} and empirically validated in Appendix~\ref{natten_ablation_study}, we restrict the zero-shot substitution to the decoder's high-resolution layers. 
The layer count $L \in \{0,1,2,3\}$ therefore denotes the number of high-resolution layers replaced with NATTEN in the decoder.

\subsection{Flow-Guided Interpolation}
\label{sec:rife}
The temporal branch fills the frames skipped by the keyframe strategy. We use RIFE \cite{huang2022rife}, a pretrained real-time interpolation model, as $\mathcal{G}_{\mathrm{Int}}$. Given two super-resolved keyframes $y_{t_{\mathrm{prev}}}$ and $y_{t_{\mathrm{next}}}$ and target timestamp $\tau_t$, RIFE predicts motion fields and a visibility mask to synthesize the intermediate frame:
\begin{equation}
    \hat{y}_t = M_t \odot \mathcal{W}(y_{t_{\mathrm{prev}}},F_{t\rightarrow \mathrm{prev}})
    + (1-M_t) \odot \mathcal{W}(y_{t_{\mathrm{next}}},F_{t\rightarrow \mathrm{next}}),
\end{equation}
where $\mathcal{W}$ denotes warping and $\odot$ denotes element-wise multiplication.

For $K=2$, the interpolator synthesizes a single midpoint frame with $\tau_t=0.5$. For $K=3$, it synthesizes two intermediate frames at $\tau_t\in\{1/3,2/3\}$. This branch is lightweight compared with the spatial generator and acts as a temporal regularizer: it suppresses flicker caused by independently generated frames, but it also introduces a predictable failure mode when motion becomes difficult for interpolation.

\subsubsection{Motion Gate.}
Frame interpolation performs very well under low-to-mid interframe motion, but it tends to break under larger motion.
We address this with a lightweight, training-free motion gate based on the peak Farneback~\cite{farneback2003} optical dense flow using a fixed threshold of $\tau_{\mathrm{gate}}=15$.
In Appendix~\ref{app:motion_gate}, we compare a set of ranges for both peak and mean optical flow magnitude.
We select $\tau_{\mathrm{gate}}=15$ as a conservative operating point that favors temporal stability.

\section{Experiments}
\label{sec:experiments}

\subsection{Experimental Setup}
\textbf{Datasets.}
We evaluate on four standard VSR benchmarks: Vid4 \cite{vid4}, REDS4 \cite{reds4}, UDM10 \cite{udm10}, and SPMCS \cite{spmcs}. All experiments use $4\times$ super-resolution.

\textbf{Baselines.}
We compare against RVRT, a recurrent transformer-based restoration model with guided deformable attention designed for balanced video restoration performance \cite{NEURIPS2022_02687e7b}; 
RealBasicVSR \cite{chan2022investigating}---a GAN-based real-world VSR model with degradation-aware cleaning, representative of the pre-diffusion perceptually-oriented regression baselines.
We also compare against DLoRAL, a one-step diffusion-based VSR method with learned low-rank adaptation modules \cite{sun2025onestepdiffusiondetailrichtemporally}; DiffIR2VR-Zero, a training-free restoration pipeline built on image diffusion priors \cite{yeh2024diffir2vr}, and lastly, 
SeedVR2 \cite{wang2025seedvr2}---a recent diffusion-based one-step VSR model, representative of the current state of the art in generative VSR quality.

\textbf{Metrics.}
We report PSNR, SSIM, and LPIPS, together with the no-reference perceptual metrics BRISQUE, MUSIQ, and MANIQA. Temporal consistency is measured using E$_{warp}$ and tOF.

\textbf{Runtime protocol.}
All runtime and memory measurements are obtained on a single NVIDIA RTX 4090 GPU. Table~\ref{tab:single_batch_performance_and_memory} single batch performance across methods. 
Table~\ref{tab:throughput_performance} then characterizes only the scalability of our own pipeline as batch size and keyframe interval vary. 

This separation reflects a fair reading of the results: the competing methods span different architectural families with distinct memory behavior, which limits the batch sizes that are feasible on the same hardware. We therefore treat batching purely as a system's property of our own pipeline.

\begin{table}
    \centering
    \footnotesize
    \caption{Throughput and memoryof the proposed pipeline at 1280x720 resolution for different keyframe intervals $K$ and batch sizes $B$.}
    \begin{tabular}{ccrr}
        \toprule
        $K$ & $B$ & Memory (GB) & FPS \\
        \midrule
        \multirow{2}{*}{1} & 1 & 2.03 & 3.04 \\
          & 8 & 8.94 & 16.76 \\
        \addlinespace
        \multirow{2}{*}{2} & 1 & 2.23 & 5.69   \\
          & 8 & 8.98 & 25.64  \\
        \addlinespace
        \multirow{2}{*}{3} & 1 & 2.33 & 8.28   \\
          & 8 & 8.99 & \textbf{32.66}  \\
        \bottomrule
    \end{tabular}
    \label{tab:throughput_performance}
\end{table}

\subsection{Efficiency}
\textbf{Single Batch.} Table~\ref{tab:single_batch_performance_and_memory} shows that even before temporal interleaving, the selected one-step diffusion backbone is substantially lighter and faster than the other generative backbones, with SeedVR2~\cite{wang2025seedvr2} at 0.97 FPS.

\textbf{Throughput.} Table~\ref{tab:throughput_performance} shows how the full pipeline scales with batch size $B$ (single video-stream) and keyframe interval $K$.
At batch size 1 -- closest to a per-frame latency operating point -- throughput reaches 3.04, 5.69, and 8.28 FPS for $K{=}1,2,3$ respectively. 
At batch size 8, throughput rises to 16.76, 25.64, and 32.66 FPS, while memory grows only marginally.

\begin{table}
    \centering
    \footnotesize
    \caption{Single-batch HD (1280x720) inference on the same GPU. Peak memory is reported by PyTorch. $L{=}3$ for all runs.}
    \begin{tabular}{lrr}
        \toprule
        Method & Memory (GB) & FPS \\
        \midrule
        \textit{Non-Diffusion} \\
        RealBasicVSR    &    4.6     &  45.53   \\
        RVRT            & 7.1 & 1.97 \\
        \addlinespace
        \textit{Diffusion-based} \\
        DLoRAL & 8.5 & 0.15 \\
        DiffIR2VR-Zero & 9.5 & 0.08 \\
        SeedVR2 (3B, FP8)             &   8.1     &   0.97   \\
        AdcSR & 2.0 & 3.04 \\
        \bottomrule
    \end{tabular}
    \label{tab:single_batch_performance_and_memory}
\end{table}

\begin{table}[ht]
    \centering
    \caption{Effect of $L{=}2$ at $K{=}1$ against baseline ($L{=}0$). Best value per dataset and metric is in \textbf{bold}.}
    \label{tab:natten_quality_summary}
    \footnotesize
    \resizebox{\linewidth}{!}{%
\begin{tabular}{ll ccccccccc}
        \toprule
        Dataset & $L$ & PSNR$\uparrow$ & SSIM$\uparrow$ & LPIPS$\downarrow$ & DISTS$\downarrow$ & BRISQUE$\downarrow$ & MUSIQ$\uparrow$ & MANIQA$\uparrow$ & $E_{\text{warp}}\downarrow$ & tOF$\downarrow$ \\
        \midrule
        \multirow{2}{*}{UDM10}
          & 0 & 21.94 & 0.7879 & \textbf{0.3731} & 0.1301 & 11.62 & \textbf{71.65} & \textbf{0.5187} & 0.0279 & 1.1653 \\
          & 2 & \textbf{23.21} & \textbf{0.8075} & 0.3733 & \textbf{0.1203} & \textbf{10.57} & 68.12 & 0.4487 & \textbf{0.0228} & \textbf{0.7843} \\
        \midrule
        \multirow{2}{*}{SPMCS}
          & 0 & 19.84 & 0.5468 & \textbf{0.4671} & \textbf{0.2006} & 14.92 & \textbf{67.75} & \textbf{0.4363} & 0.0364 & 1.1553 \\
          & 2 & \textbf{20.33} & \textbf{0.5443} & 0.4793 & 0.2288 & \textbf{13.62} & 63.96 & 0.3831 & \textbf{0.0315} & \textbf{0.9226} \\
        \midrule
        \multirow{2}{*}{Vid4}
          & 0 & 18.17 & 0.6116 & 0.4290 & \textbf{0.1535} & \textbf{6.50} & \textbf{70.74} & \textbf{0.4397} & 0.0432 & \textbf{0.2589} \\
          & 2 & \textbf{18.92} & \textbf{0.6161} & \textbf{0.4288} & 0.1541 & 7.42 & 68.73 & 0.3936 & \textbf{0.0387} & 0.2754 \\
        \midrule
        \multirow{2}{*}{REDS4}
          & 0 & 20.92 & 0.7942 & \textbf{0.3596} & 0.1120 & \textbf{11.41} & \textbf{71.76} & \textbf{0.4893} & 0.0391 & 0.8986 \\
          & 2 & \textbf{21.59} & \textbf{0.8058} & 0.3664 & \textbf{0.1074} & 12.76 & 69.23 & 0.4520 & \textbf{0.0349} & \textbf{0.8184} \\
        \bottomrule
    \end{tabular}}
\end{table}

\subsection{Effect of Zero-Shot Attention Substitution}

Table~\ref{tab:natten_quality_summary} compares the $L{=}2$ against the baseline $L{=}0$ at $K{=}1$. 
We examine reconstruction quality, temporal stability, and inference speed in turn.
A more extensive study across $L$ and $K$ is available in Appendix~\ref{natten_ablation_study}.

\textbf{Quality.}
The $L{=}2$ improves PSNR and SSIM across all datasets, LPIPS remains stable to slightly increasing. 
Results for DISTS and BRISQUE are mixed; neither configuration is preferred.
MUSIQ and MANIQA, however, consistently worsen compared to the baseline.

\textbf{Temporal Stability.}
Ewarp is a clear improvement across all datasets when $L{=}2$, and tOF shows substantial improvement in UDM10 (\textit{-0.3810}) and SPCMS (\textit{0.2327}), a smaller improvement in REDS4 (\textit{-0.0802}), and a very slight deterioration on Vid4 (\textit{+0.0165}).

\textbf{Speed.}
Table~\ref{tab:natten_speed} shows how batch size and resolution affect the speed gains NATTEN can provide, culminating in up to 44.7\% speedup at 4K. 
The speed increase is linear with respect to the number of $L$.

\begin{table}[t]
    \centering
    \footnotesize
    \setlength{\tabcolsep}{4pt}
    \caption{Inference speed with NATTEN relative to PyTorch FlashAttention for $K{=}1$ and $L{=}3$.}
    \begin{tabular}{rcrrr}
        \toprule
        Resolution & B & FA FPS & NATTEN FPS & Change \\
        \midrule
        1280$\times$720 & 1 & 3.01 & 2.98 & -1.0\% \\
                         & 8 & 15.30 & 16.76 & +9.1\% \\
        1920$\times$1080 & 1 & 3.06 & 3.00 & -1.8\% \\
                          & 4 & 6.20 & 7.31 & +18.0\% \\
        2560$\times$1440 & 1 & 2.73 & 3.49 & +27.9\% \\
        3840$\times$2160 & 1 & 0.76 & 1.10 & +44.7\% \\
        \bottomrule
    \end{tabular}
    \label{tab:natten_speed}
\end{table}

\subsection{Quality on Standard Benchmarks} \label{Quality on Standard Benchmarks}

\begin{table}[t]
    \centering
    \small
    \setlength{\tabcolsep}{4pt}
    \renewcommand{\arraystretch}{1.0}
    \caption{Quantitative comparison on UDM10, SPMCS, REDS4, and Vid4 for $\times 4$ super-resolution. Best results are marked in \textbf{bold}, second-best in \textit{\textcolor{secondbest}{italic green}}. $K$ is the interpolation interval, $L$ are the attention layers replaced with NATTEN.}
    \resizebox{\linewidth}{!}{%
\begin{tabular}{clrrrrrrrr}
        \toprule
        & & & & & & & \multicolumn{3}{c}{\textbf{Ours}} \\
        & \textbf{Metric} & \textbf{RealBasicVSR} & \textbf{RVRT} & 
        \textbf{DLoRAL} & \textbf{DiffIR2VR-Zero} & \textbf{SeedVR2} & 
        $K=1$, $L=2$ & $K=2$, $L=2$ & $K=3$, $L=2$ \\
        \midrule
        \multirow{8}{*}{\rotatebox{90}{\textbf{UDM10 $\times$4}}}
          & PSNR $\uparrow$               & 29.389 & \textbf{33.279} & 20.957 & 24.929 & \ssb{29.470} & 23.206 & 23.208 & 23.111 \\
          & SSIM $\uparrow$               & 0.937 & \textbf{0.985} & 0.694 & 0.854 & \ssb{0.939} & 0.807 & 0.805 & 0.801 \\
          & LPIPS $\downarrow$            & 0.253 & \textbf{0.154} & 0.325 & 0.389 & \ssb{0.214} & 0.373 & 0.368 & 0.368 \\
          & BRISQUE $\downarrow$          & 21.295 & 49.080 & 29.212 & 22.888 & 25.090 & \textbf{10.569} & \ssb{13.694} & 13.767 \\
          & MUSIQ $\uparrow$              & 65.784 & 59.387 & 62.180 & 67.046 & 65.900 & \textbf{68.123} & \ssb{67.676} & 67.555 \\
          & MANIQA $\uparrow$             & 0.353 & 0.307 & 0.350 & \textbf{0.525} & 0.372 & \ssb{0.449} & 0.437 & 0.432 \\
          & E$_{warp}$ $\downarrow$       & \ssb{0.010} & \textbf{0.006} & 0.019 & 0.018  & 0.017 & 0.023 & 0.016 & 0.015 \\
          & tOF $\downarrow$              & \ssb{0.314} & \textbf{0.127} & 0.659 & 0.511 & 0.424 & 0.784 & 0.601 & 0.875 \\
        \midrule
        \multirow{8}{*}{\rotatebox{90}{\textbf{SPMCS $\times$4}}}
          & PSNR $\uparrow$               & \textbf{22.292} & \ssb{22.233} & 21.182 & 21.792 & 22.030 & 20.334 & 20.452 & 20.449 \\
          & SSIM $\uparrow$               & \ssb{0.637} & 0.597 & 0.625 & \textbf{0.661} & 0.636 & 0.544 & 0.551 & 0.550 \\
          & LPIPS $\downarrow$            & 0.445 & 0.539 & \textbf{0.325} & \ssb{0.405} & 0.413 & 0.479 & 0.476 & 0.475 \\
          & BRISQUE $\downarrow$          & \textbf{13.435} & 66.396 & 20.185 & 24.269 & 16.420 & \ssb{13.621} & 15.218 & 14.961 \\
          & MUSIQ $\uparrow$              & 67.245 & 25.621 & \ssb{67.501} & \textbf{68.980} & 65.390 & 63.964 & 63.110 & 63.256 \\
          & MANIQA $\uparrow$             & 0.334 & 0.193 & \ssb{0.418} & \textbf{0.545} & 0.407 & 0.383 & 0.373 & 0.372 \\
          & E$_{warp}$ $\downarrow$       & \ssb{0.018} & 0.020 & 0.021 & 0.028 & 0.031 & 0.032 & 0.021 & \textbf{0.017} \\
          & tOF $\downarrow$              & \textbf{0.361} & 2.993 & 0.547 & 0.664 & 0.602 & 0.923 & 0.420 & \ssb{0.391} \\
        \midrule
        \multirow{8}{*}{\rotatebox{90}{\textbf{REDS4 $\times$4}}}
          & PSNR $\uparrow$               & \ssb{27.042} & \textbf{32.480} & 19.162 & 25.142 & 25.250 & 21.593 & 20.440 & 19.570 \\
          & SSIM $\uparrow$               & \ssb{0.945} & \textbf{0.990} & 0.459 & 0.885 & 0.913 & 0.806 & 0.713 & 0.649 \\
          & LPIPS $\downarrow$            & \ssb{0.231} & \textbf{0.178} & 0.451 & 0.351 & 0.252 & 0.366 & 0.390 & 0.410 \\
          & BRISQUE $\downarrow$          & 18.052 & 38.871 & 13.680 & \ssb{13.587} & 16.500 & \textbf{12.762} & 16.080 & 15.920\\
          & MUSIQ $\uparrow$              & 67.034 & 64.661 & 65.367 & \textbf{70.606} & \ssb{69.540} & 69.235 & 68.190 & 67.920 \\
          & MANIQA $\uparrow$             & 0.354 & 0.355 & 0.397 & \textbf{0.606} & 0.433 & \ssb{0.452} & 0.427 & 0.418 \\
          & E$_{warp}$ $\downarrow$       & \ssb{0.019} & \textbf{0.011} & 0.064 & 0.029 & 0.034 & 0.035 & 0.051 & 0.056 \\
          & tOF $\downarrow$              & \ssb{0.510} & \textbf{0.299} & 6.785 & 0.644 & 0.574 & 0.818 & 4.088 & 5.226 \\
        \midrule
        \multirow{8}{*}{\rotatebox{90}{\textbf{Vid4 $\times$4}}}
          & PSNR $\uparrow$               & \ssb{22.834} & \textbf{26.552} & 18.803 & 21.394 & 21.080 & 18.917 & 18.968 & 18.886 \\
          & SSIM $\uparrow$               & \ssb{0.823} & \textbf{0.938} & 0.453 & 0.708 & 0.757 & 0.616 & 0.619 & 0.613 \\
          & LPIPS $\downarrow$            & \ssb{0.312} & \textbf{0.234} & 0.513 & 0.335  & 0.323 & 0.429 & 0.429 & 0.430 \\
          & BRISQUE $\downarrow$          & 21.895 & 38.672 & 18.500 & 24.093 & 19.700 & \textbf{7.421} & 9.303 & \ssb{9.050} \\
          & MUSIQ $\uparrow$              & 67.367 & 58.221 & 59.648 & \textbf{69.847} & \ssb{69.350} & 68.731 & 67.892 & 67.803 \\
          & MANIQA $\uparrow$             & 0.310 & 0.279 & 0.301 & \textbf{0.480} & 0.381 & \ssb{0.394} & 0.380 & 0.377 \\
          & E$_{warp}$ $\downarrow$       & \ssb{0.023} & \textbf{0.011} & 0.029 & 0.042 & 0.041 & 0.039 & 0.028 & 0.024 \\
          & tOF $\downarrow$              & \ssb{0.081} & \textbf{0.049} & 0.256 & 0.189 & 0.184 & 0.275 & 0.218 & 0.267 \\
        \bottomrule
    \end{tabular}}
    \label{tab:comparison_with_sota}
\end{table}

Table~\ref{tab:comparison_with_sota} summarizes the quality trade-offs on UDM10, SPMCS, REDS4, and Vid4 against RealBasicVSR, RVRT, DLoRAL, DiffIR2VR-Zero, and SeedVR2.

\textbf{Quality.} 
Our approach occupies a favorable perceptual-efficiency regime: it achieves the best BRISQUE values on three of four datasets and ranks first or second.
It places third on MUSIQ on REDS4 and Vid4, but falls to fifth of eight on SPMCS, behind four baselines.
In reference-based perceptual metrics such as LPIPS, our approach typically falls between DLoRAL and DiffIR2VR-Zero, except on SPMCS, where it trails both. 
REDS4 is the hardest case, where perceptual metrics stay competitive, but PSNR, SSIM, and LPIPS degrade as $K$ increases, exposing the limits of relying on interpolation rather than jointly modeling temporal generation under complex motion.

\begin{table}[ht]
  \centering
  \caption{Super-resolution quality and temporal-consistency metrics on REDS4 under the adaptive motion-gated interpolation schedule with $L{=}2$ and key-frame intervals $K \in \{2,3\}$. Fallback rates for $K{=}2$ are 74\% and  90\% for $K{=}3$.}
  \footnotesize
  \resizebox{\linewidth}{!}{%
\begin{tabular}{l cccccc ccc}
    \toprule
    $K$ & PSNR$\uparrow$ & SSIM$\uparrow$ & LPIPS$\downarrow$ & 
    BRISQUE$\downarrow$ & MUSIQ$\uparrow$ & MANIQA$\uparrow$ &
    $E_{\text{warp}}\downarrow$ & tOF$\downarrow$ & LPIPS$_{\text{inter}}\downarrow$ \\
    \midrule
    2 (Ungated) & 20.440 & 0.713 & 0.390 & 
    16.080 & 68.190 & 0.427 & 0.051 & 4.088  & \textbf{0.364} \\
    2 (Gated) & \textbf{21.500} & \textbf{0.796} & \textbf{0.368} & 
    \textbf{13.410} & \textbf{68.990} & \textbf{0.445} & \textbf{0.034} & \textbf{1.079} & 0.370 \\
    
    \addlinespace
    3 (Ungated) & 19.570 & 0.649 & 0.410 & 
    15.920 & 67.920 & 0.418 & 0.056 & 5.226  & \textbf{0.364} \\
    
    3 (Gated) & \textbf{21.570} & \textbf{0.804} & \textbf{0.367} &
    \textbf{12.950} & \textbf{69.120} & \textbf{0.449} & \textbf{0.034} & \textbf{0.848} & 0.372 \\
    \bottomrule
    
    \end{tabular}}
    \label{tab:motion_gate_on_reds4}
\end{table}

\textbf{Temporal consistency.}
E$_{\mathrm{warp}}$ and tOF tell a related but distinct story.
On SPMCS, both metrics improve monotonically as $K$ increases from 1 to 3.
On UDM10 and Vid4, E$_{\mathrm{warp}}$ improves consistently with $K$, and tOF improves at $K{=}2$; on Vid4 this holds through $K{=}3$, but on UDM10 it turns back upward at $K{=}3$ (0.875).
In REDS4, both E$_{\mathrm{warp}}$ and tOF worsen from $K{=}1$ to $K{=}2$, since its large inter-frame motion falls outside the range RIFE can interpolate reliably -- which is why we introduce the motion gate.
Section~\ref{sec:motion_gating} shows that this degradation can be substantially mitigated with a simple motion-adaptive gate, at the cost of throughput (see Table~\ref{tab:motion_gate_on_reds4}).

\subsection{Motion-Gated Interpolation on High-Motion Content}
\label{sec:motion_gating}

Section~\ref{Quality on Standard Benchmarks} showed that tOF degrades sharply on REDS4 as $K$ increases, a
consequence of frame interpolation breaking down under REDS4's large inter-frame motion. 

Table~\ref{tab:motion_gate_on_reds4} reports quality and temporal-consistency metrics on REDS4 under this gated schedule for $K \in \{2,3\}$. 
The effect on temporal consistency is substantial: at $L{=}2$, gating reduces tOF from 4.088 to 1.079 at $K{=}2$ (a 74\% reduction) and from 5.226 to 0.848 at $K{=}3$ (an 84\% reduction), bringing both configurations close to the $K{=}1$ reference value of 0.818. $E_{\text{warp}}$  and PSNR/SSIM revert to being much closer to the $K{=}1$ configuration.

The throughput cost when using the motion gate on REDS4 is quite impactful. 
At $K{=}2$, the fallback rate of $74\%$ limits throughput to $18.33$ FPS for a +9.4\% against the ungated +54.6\% gain.
For $K{=}3$, the fallback rate is higher at $90\%$, limiting throughput to $17.37$ FPS, just +3.5\% against the ungated +94.9\% gain.


On lower motion content, the throughput cost is much lower. 
On UDM10, it activates in 20\% of the total frames; for SPMCS and Vid4, it activates in 2.6\% and 7\%, respectively (scaled to match the resolution of the REDS4 frames). 

\subsection{Scope, Limitations, and Reading of the Claims}
\label{sec:limitations}
Higher throughput in our pipeline comes at the cost of higher latency: batching up to $B{=}8$ introduces up to $B \cdot K$ frames of buffering latency (e.g., 24 frames, roughly 0.8--1\,s at 24--30\,fps, for $B{=}8$, $K{=}3$) before a batch's outputs are emitted, making this configuration suited to offline, bulk, or delay-tolerant streaming deployment (e.g., archival transcoding, batch export, broadcast with delay) rather than interactive low-latency streaming.



\subsection{Practical Implications}
The proposed design also suggests a broader lesson for efficient multimedia generation. 
Once strong image diffusion priors become available in compressed one-step form, system-level choices such as scheduling, attention placement, and module composition can become at least as important as designing new monolithic architectures. 
The results here indicate that a considerable portion of the speed-quality frontier can be captured by carefully orchestrating pretrained components, provided that the trade-offs are made explicit.

\section{Conclusion}
We present a modular zero-shot framework for generative video super-resolution built entirely from pretrained components. 
By decoupling spatial enhancement---via a one-step diffusion model---from temporal stabilization---via a lightweight, motion-gated frame interpolator---and by zero-shot substituting targeted global self-attention layers with neighborhood attention, we achieve high-throughput generative video upscaling on commodity hardware, up to 32.66 FPS at 720p on a single RTX 4090, trading reconstruction fidelity for perceptual quality and temporal stability.


\section*{Acknowledgements}
This work was funded by the European Union's Horizon Europe research and innovation programme under the Marie Sk\l{}odowska-Curie grant agreement No.~101129865 (ACCESS).

\bibliographystyle{plainnat}
\bibliography{references}


\clearpage
\appendix

\section{NATTEN Ablation Study} \label{natten_ablation_study}

\paragraph{Replacing Global Attention with NATTEN in the UNet.}
Table~\ref{tab:down_vs_up_k1} isolates \emph{where} in the UNet the NATTEN substitution should be applied, comparing encoder-side (down-sampling) against decoder-side (up-sampling) replacement at matched layer counts $L \in \{1, 2\}$ on the full UDM10 dataset. 
Replacing attention in the up-sampling stage is consistently preferable: it achieves the best PSNR, SSIM, LPIPS, DISTS, MANIQA, and $E_{\text{warp}}$ at both $L{=}1$ and $L{=}2$, while down-sampling replacement wins on tOF, and, at $L{=}1$, on BRISQUE and MUSIQ by a very small margin. 
The gap widens rather than narrows as more layers are replaced: at $L{=}2$, down-sampling replacement trails up-sampling by 0.89\,dB in PSNR (22.32 vs.\ 23.21) and 0.0347 in SSIM (0.7728 vs.\ 0.8075), and its DISTS and BRISQUE actually regress \emph{past} the $L{=}0$ baseline (0.1457 and 12.82 vs.\ 0.1301 and 11.62), indicating that aggressively localizing attention in the encoder actively harms reconstruction quality rather than merely offering a smaller improvement than up-sampling replacement.

This asymmetry is consistent with the differing roles of the two stages. 
The encoder compresses the input into a semantic bottleneck, where attention likely aggregates global context that later stages depend on. 
Restricting it to local context neighborhoods removes information that the rest of the network cannot recover downstream. 
The decoder, by contrast, operates on features that are already semantically resolved and is primarily responsible for spatial detail synthesis, where attention plausibly captures more local, texture-level dependencies that neighborhood attention can approximate without loss --- and, at $L{=}2$, without even sacrificing fidelity relative to global attention. 
Together, these results justify our design choice to restrict the zero-shot substitution to the up-sampling path throughout the paper.

\begin{table}[H]
  \centering
  \caption{NATTEN attention replacement in the UNet's DownSampling (encoder) vs.\ UpSampling (decoder) stage, head-to-head at matched layer count $L$, on the full UDM10 dataset (10 clips) with $K{=}1$.}
  \footnotesize
  \resizebox{\linewidth}{!}{%
\begin{tabular}{cl ccccccc cc}
    \toprule
    $L$ & stage & PSNR$\uparrow$ & SSIM$\uparrow$ & LPIPS$\downarrow$ & DISTS$\downarrow$ &
    BRISQUE$\downarrow$ & MUSIQ$\uparrow$ & MANIQA$\uparrow$ &
    $E_{\text{warp}}\downarrow$ & tOF$\downarrow$ \\
    \midrule
    0 & baseline & 21.9362 & 0.7879 & 0.3731 & 0.1301 & 11.6197 & 71.6498 & 0.5187 & 0.0279 & 1.1653 \\
    \midrule
    \multirow{2}{*}{1}
      & up   & \textbf{22.4926} & \textbf{0.7995} & \textbf{0.3705} & \textbf{0.1237} & 11.4900 & 70.3318 & \textbf{0.4915} & \textbf{0.0263} & 1.0258 \\
      & down & 22.0091 & 0.7777 & 0.3883 & 0.1342 & \textbf{11.3235} & \textbf{70.5214} & 0.4881 & 0.0266 & \textbf{0.7196} \\
    \midrule
    \multirow{2}{*}{2}
      & up   & \textbf{23.2063} & \textbf{0.8075} & \textbf{0.3733} & \textbf{0.1203} & \textbf{10.5693} & 68.1227 & \textbf{0.4487} & \textbf{0.0228} & 0.7843 \\
      & down & 22.3193 & 0.7728 & 0.4028 & 0.1457 & 12.8239 & \textbf{68.5334} & 0.4450 & 0.0245 & \textbf{0.6884} \\
    \bottomrule
  \end{tabular}}
  \label{tab:down_vs_up_k1}
\end{table}

\begin{figure}[H]
  \centering
  \includegraphics[width=0.6\linewidth]{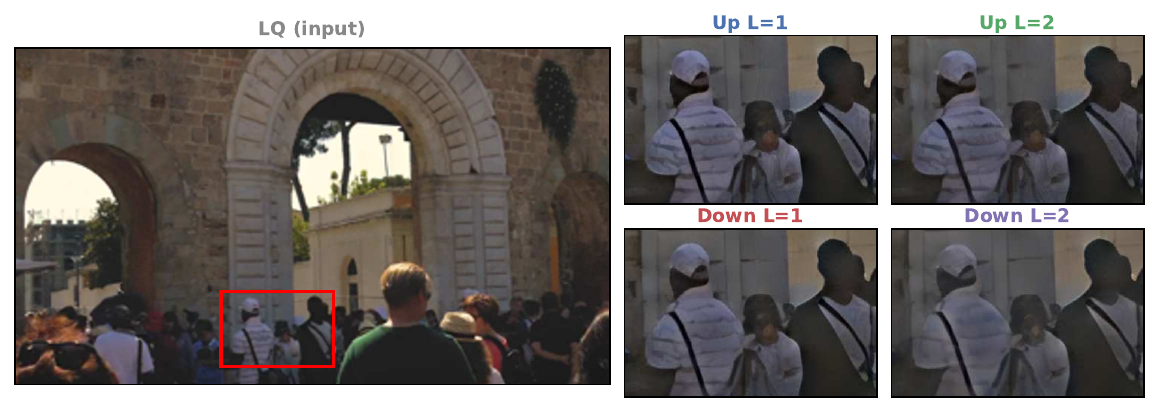}
  \caption{UDM10 clip \texttt{000}: comparison on replacing NATTEN on encoder vs decoder of UNet.}
  \label{fig:udm10_000_encoder_vs_decoder_natten_comparison}

  \vspace{2mm}
  \includegraphics[width=0.6\linewidth]{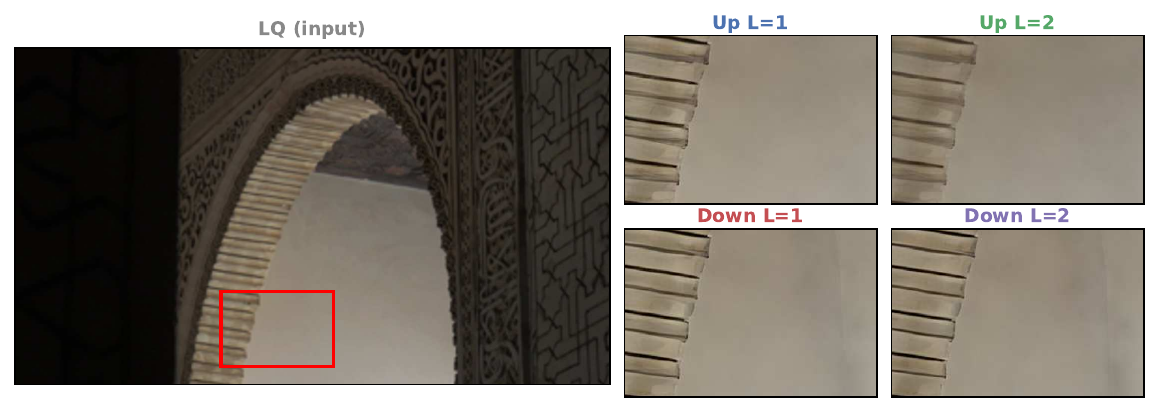}
  \caption{UDM10 clip \texttt{001}: comparison on replacing NATTEN on encoder vs decoder of UNet.}
  \label{fig:udm10_001_encoder_vs_decoder_natten_comparison}
\end{figure}

\begin{table}[!ht]
\centering
\caption{Ablation on the number of NATTEN-replaced attention layers ($L$) on UDM10, across $K$. Best value per $K$ group and metric is in \textbf{bold}.
$\uparrow$ ($\downarrow$) indicates that higher (lower) is better.}
\label{tab:natten_ablation}
\footnotesize
\setlength{\tabcolsep}{4pt}

\resizebox{\linewidth}{!}{%
\begin{tabular}{ll c cccccccccc}
  \toprule
  Dataset & $K$ & $L$ & PSNR$\uparrow$ & SSIM$\uparrow$ & LPIPS$\downarrow$ & DISTS$\downarrow$ & BRISQUE$\downarrow$ & MUSIQ$\uparrow$ & MANIQA$\uparrow$ &
LPIPS$_{\text{inter}}\!\downarrow$ & $E_{\text{warp}}\downarrow$ & tOF$\downarrow$ \\
  \midrule
  \multirow{13}{*}{UDM10}
    & \multirow{4}{*}{1}
      & 0 & 21.94 & 0.7879 & 0.3731 & 0.1301 & 11.62 & \textbf{71.65} & \textbf{0.5187} & \textbf{0.2155} & 0.0279 & 1.1653 \\
    & & 1 & 22.49 & 0.7995 & \textbf{0.3705} & 0.1237 & 11.49 & 70.33 & 0.4915 & 0.2190 & 0.0263 & 1.0258 \\
    & & 2 & 23.21 & 0.8075 & 0.3733 & \textbf{0.1203} & \textbf{10.57} & 68.12 & 0.4487 & 0.2200 & \textbf{0.0228} & \textbf{0.7843} \\
    & & 3 & \textbf{23.29} & \textbf{0.8085} & 0.3806 & 0.1263 & 12.80 & 68.17 & 0.4518 & 0.2209 & 0.0245 & 0.9951 \\
    \cmidrule(l){2-13}
    & \multirow{4}{*}{2}
      & 0 & 21.97 & 0.7862 & 0.3689 & 0.1297 & 14.33 & \textbf{71.49} & \textbf{0.5095} & \textbf{0.2003} & 0.0205 & 0.6811 \\
    & & 1 & 22.52 & 0.7976 & \textbf{0.3660} & 0.1231 & 14.36 & 70.08 & 0.4813 & 0.2033 & 0.0190 & 0.6257 \\
    & & 2 & 23.21 & 0.8054 & 0.3681 & \textbf{0.1196} & \textbf{13.69} & 67.68 & 0.4366 & 0.2058 & \textbf{0.0163} & \textbf{0.6006} \\
    & & 3 & \textbf{23.30} & \textbf{0.8066} & 0.3748 & 0.1258 & 15.75 & 67.78 & 0.4400 & 0.2079 & 0.0176 & 0.7467 \\
    \cmidrule(l){2-13}
    & \multirow{4}{*}{3}
      & 0 & 21.87 & 0.7797 & 0.3691 & 0.1304 & 14.54 & \textbf{71.37} & \textbf{0.5057} & \textbf{0.1806} & 0.0194 & 1.1075 \\
    & & 1 & 22.42 & 0.7919 & \textbf{0.3660} & 0.1239 & 14.50 & 69.97 & 0.4776 & 0.1819 & 0.0174 & 0.9589 \\
    & & 2 & 23.11 & 0.8008 & 0.3678 & \textbf{0.1204} & \textbf{13.77} & 67.56 & 0.4325 & 0.1816 & \textbf{0.0145} & \textbf{0.8749} \\
    & & 3 & \textbf{23.19} & \textbf{0.8019} & 0.3747 & 0.1261 & 15.76 & 67.64 & 0.4360 & 0.1847 & 0.0157 & 0.9941 \\
  \midrule
  \multirow{13}{*}{SPMCS}
    & \multirow{4}{*}{1}
      & 0 & 19.84 & 0.5468 & \textbf{0.4671} & \textbf{0.2006} & 14.92 & \textbf{67.75} & \textbf{0.4363} & \textbf{0.2116} & 0.0364 & 1.1553 \\
    & & 1 & 20.13 & 0.5469 & 0.4700 & 0.2096 & 14.43 & 66.34 & 0.4128 & 0.2169 & 0.0348 & 1.0739 \\
    & & 2 & \textbf{20.33} & 0.5443 & 0.4793 & 0.2288 & \textbf{13.62} & 63.96 & 0.3831 & 0.2175 & \textbf{0.0315} & \textbf{0.9226} \\
    & & 3 & 19.93 & \textbf{0.5498} & 0.4801 & 0.2246 & 17.52 & 64.04 & 0.3707 & 0.2154 & 0.0333 & 1.1517 \\
    \cmidrule(l){2-13}
    & \multirow{4}{*}{2}
      & 0 & 20.00 & 0.5545 & \textbf{0.4632} & \textbf{0.2013} & 16.22 & \textbf{67.23} & \textbf{0.4284} & \textbf{0.1603} & 0.0245 & 0.4266 \\
    & & 1 & 20.28 & 0.5543 & 0.4661 & 0.2101 & 15.79 & 65.67 & 0.4043 & 0.1640 & 0.0232 & 0.4227 \\
    & & 2 & \textbf{20.45} & 0.5509 & 0.4756 & 0.2295 & \textbf{15.22} & 63.11 & 0.3734 & 0.1654 & \textbf{0.0208} & 0.4203 \\
    & & 3 & 20.05 & \textbf{0.5574} & 0.4762 & 0.2251 & 19.45 & 63.23 & 0.3621 & 0.1657 & 0.0219 & \textbf{0.4181} \\
    \cmidrule(l){2-13}
    & \multirow{4}{*}{3}
      & 0 & 20.00 & 0.5537 & \textbf{0.4626} & \textbf{0.2005} & 16.05 & \textbf{67.37} & \textbf{0.4271} & \textbf{0.1263} & 0.0201 & 0.3978 \\
    & & 1 & 20.27 & 0.5534 & 0.4655 & 0.2093 & 15.56 & 65.83 & 0.4028 & 0.1283 & 0.0190 & 0.3945 \\
    & & 2 & \textbf{20.45} & 0.5501 & 0.4748 & 0.2288 & \textbf{14.96} & 63.26 & 0.3718 & 0.1278 & \textbf{0.0170} & 0.3908 \\
    & & 3 & 20.05 & \textbf{0.5567} & 0.4754 & 0.2243 & 19.15 & 63.33 & 0.3603 & 0.1283 & 0.0179 & \textbf{0.3900} \\
  \midrule
  \multirow{13}{*}{Vid4}
    & \multirow{4}{*}{1}
      & 0 & 18.17 & 0.6116 & 0.4290 & 0.1535 & \textbf{6.50} & \textbf{70.74} & \textbf{0.4397} & \textbf{0.1833} & 0.0432 & \textbf{0.2589} \\
    & & 1 & 18.51 & 0.6171 & \textbf{0.4277} & \textbf{0.1486} & 7.05 & 69.83 & 0.4188 & 0.1869 & 0.0417 & 0.2649 \\
    & & 2 & \textbf{18.92} & 0.6161 & 0.4288 & 0.1541 & 7.42 & 68.73 & 0.3936 & 0.1891 & \textbf{0.0387} & 0.2754 \\
    & & 3 & 18.88 & \textbf{0.6244} & 0.4321 & 0.1553 & 10.54 & 67.38 & 0.3844 & 0.1891 & 0.0414 & 0.2660 \\
    \cmidrule(l){2-13}
    & \multirow{4}{*}{2}
      & 0 & 18.26 & 0.6154 & 0.4287 & 0.1555 & \textbf{8.26} & \textbf{70.22} & \textbf{0.4300} & \textbf{0.1688} & 0.0318 & 0.2158 \\
    & & 1 & 18.59 & 0.6208 & \textbf{0.4275} & \textbf{0.1507} & 8.86 & 69.22 & 0.4080 & 0.1720 & 0.0305 & 0.2160 \\
    & & 2 & \textbf{18.97} & 0.6187 & 0.4290 & 0.1563 & 9.30 & 67.89 & 0.3802 & 0.1743 & \textbf{0.0281} & 0.2182 \\
    & & 3 & 18.95 & \textbf{0.6286} & 0.4324 & 0.1579 & 12.90 & 66.43 & 0.3738 & 0.1755 & 0.0301 & \textbf{0.2122} \\
    \cmidrule(l){2-13}
    & \multirow{4}{*}{3}
      & 0 & 18.18 & 0.6096 & 0.4302 & 0.1566 & \textbf{7.98} & \textbf{70.15} & \textbf{0.4256} & \textbf{0.1467} & 0.0274 & 0.2715 \\
    & & 1 & 18.50 & 0.6149 & \textbf{0.4289} & \textbf{0.1515} & 8.56 & 69.22 & 0.4044 & 0.1483 & 0.0260 & 0.2699 \\
    & & 2 & \textbf{18.89} & 0.6132 & 0.4300 & 0.1567 & 9.05 & 67.80 & 0.3766 & 0.1489 & \textbf{0.0238} & 0.2671 \\
    & & 3 & 18.87 & \textbf{0.6230} & 0.4332 & 0.1579 & 12.72 & 66.38 & 0.3695 & 0.1504 & 0.0255 & \textbf{0.2600} \\
  \midrule
  \multirow{13}{*}{REDS4}
    & \multirow{4}{*}{1}
      & 0 & 20.92 & 0.7942 & \textbf{0.3596} & 0.1120 & \textbf{11.41} & \textbf{71.76} & \textbf{0.4893} & 0.3803 & 0.0391 & 0.8986 \\
    & & 1 & 21.30 & 0.8028 & 0.3598 & \textbf{0.1073} & 12.45 & 70.65 & 0.4737 & 0.3793 & 0.0379 & 0.8413 \\
    & & 2 & \textbf{21.59} & \textbf{0.8058} & 0.3664 & 0.1074 & 12.76 & 69.23 & 0.4520 & \textbf{0.3756} & \textbf{0.0349} & \textbf{0.8184} \\
    & & 3 & 21.39 & 0.8008 & 0.3683 & 0.1141 & 17.46 & 69.47 & 0.4506 & 0.3779 & 0.0377 & 0.8238 \\
    \cmidrule(l){2-13}
    & \multirow{4}{*}{2}
      & 0 & 19.73 & 0.6983 & \textbf{0.3847} & 0.1171 & \textbf{13.98} & \textbf{71.23} & \textbf{0.4733} & 0.3673 & 0.0604 & 4.0879 \\
    & & 1 & 20.09 & 0.7068 & \textbf{0.3847} & 0.1127 & 15.30 & 69.91 & 0.4546 & 0.3666 & 0.0573 & 4.0876 \\
    & & 2 & \textbf{20.44} & \textbf{0.7133} & 0.3904 & \textbf{0.1126} & 16.08 & 68.19 & 0.4273 & \textbf{0.3640} & \textbf{0.0510} & 4.0881 \\
    & & 3 & 20.21 & 0.7041 & 0.3921 & 0.1194 & 20.83 & 68.45 & 0.4278 & 0.3668 & 0.0563 & \textbf{4.0670} \\
    \cmidrule(l){2-13}
    & \multirow{4}{*}{3}
      & 0 & 18.85 & 0.6314 & 0.4056 & 0.1203 & \textbf{13.99} & \textbf{71.04} & \textbf{0.4648} & 0.3503 & 0.0660 & 5.3189 \\
    & & 1 & 19.20 & 0.6405 & \textbf{0.4050} & \textbf{0.1158} & 15.24 & 69.66 & 0.4462 & 0.3489 & 0.0624 & 5.2732 \\
    & & 2 & \textbf{19.57} & \textbf{0.6490} & 0.4097 & 0.1160 & 15.92 & 67.92 & 0.4183 & \textbf{0.3455} & \textbf{0.0556} & 5.2260 \\
    & & 3 & 19.32 & 0.6383 & 0.4113 & 0.1225 & 20.67 & 68.21 & 0.4194 & 0.3487 & 0.0610 & \textbf{5.2143} \\
    
  \bottomrule
\end{tabular}}
\end{table}

\paragraph{Effect of NATTEN layer count.} Table~\ref{tab:natten_ablation} reports the effect of replacing $L \in \{0,1,2,3\}$ attention layers of the upsampling branch with NATTEN local attention, evaluated on four standard VSR benchmarks (UDM10, SPMCS, Vid4, REDS4) across three keyframe intervals ($K \in \{1,2,3\}$). 
Two trends emerge consistently across all datasets and $K$ settings.

\paragraph{Quality.}
First, pixel-fidelity metrics improve as more layers are replaced: PSNR increases monotonically from $L{=}0$ to $L{=}2$ in all four datasets (e.g., PSNR gains of up to $+1.27$ dB on UDM10 and $+0.67$ dB on REDS4 at $K{=}1$), and either continue improving slightly ($L{=}3$ on UDM10) or plateau/regress marginally at $L{=}3$.
In a similar manner, SSIM seems to improve across all datasets up to $L{=}3$, with the exception of REDS4, peaking at $L{=}2$.
This indicates that local attention primarily aids reconstruction fidelity, with diminishing or even negative returns once more than two layers are replaced.
Second, this fidelity gain does not transfer uniformly to perceptual quality. 
LPIPS and DISTS are consistently best at low-to-moderate $L$ (0--2, never at $L{=}3$). 
At the same time, the no-reference metrics MUSIQ and MANIQA are maximized at $L{=}0$ in \emph{every} dataset/$K$ combination we tested.
BRISQUE is the lowest at $L{=}2$ on UDM10 and SPMCS, and the lowest on $L{=}0$ on Vid4 and REDS4.

\paragraph{Temporal consistency.}
The warping error $E_{\text{warp}}$ shows the cleanest signal in the whole study: it is minimized at $L{=}2$ for all four datasets and all three $K$ values without exception, followed by $L{=}3$ and $L{=}1$, always lower than the $L{=}0$ baseline.
LPIPS$_{\text{inter}}$ (inter-frame perceptual distance) favors $L{=}0$ on three of the four datasets. Still, this metric should be interpreted cautiously, as a lower value can also reflect over-smoothed, lower-detail output rather than genuinely improved temporal stability.
tOF follows a similar direction to $E_{\text{warp}}$ (favoring $L{=}2$--$3$) on UDM10/SPMCS and Vid4, but rises sharply on REDS4 as $K$ increases (from $\sim$0.8 at $K{=}1$ to over 5.0 at $K{=}3$); we attribute this to REDS4's characteristically large inter-frame motion interacting poorly with frame interpolation, since wider keyframe gaps force the interpolation module to synthesize intermediate frames over larger displacements than it can reliably track, producing optical-flow-inconsistent frames that tOF penalizes heavily.

\paragraph{Recommendation.} Taken together, $L{=}2$ offers the best overall balance in our ablation: it nearly matches the peak PSNR/SSIM achieved at $L{=}3$, retains competitive LPIPS/DISTS, and uniquely minimizes $E_{\text{warp}}$ across every dataset and $K$ setting, at the cost of a modest drop in no-reference quality (MUSIQ/MANIQA) relative to the unmodified ($L{=}0$) baseline.
Similarly, tOF is almost always the lowest at $L{=}2$ or $L{=}3$, with the only exception being the $K{=}1$ setting for SPMCS.

As for $K$, it is dependent on the amount of movement across frames. In low movement (UDM10, SPMCS), $K{=}3$ is acceptable; for moderate movement (Vid4), it shows slight degradation relative to $K{=}2$; and for large movement (REDS4), even $K{=}2$ shows a large increase in tOF.

\FloatBarrier
\subsection{Same frames, different L}
\raggedbottom
\begingroup
\setlength{\intextsep}{6pt plus 2pt minus 2pt}
\setlength{\abovecaptionskip}{-6pt}
\setlength{\belowcaptionskip}{0pt}
\begin{figure}[H]
  \centering
  \includegraphics[width=\linewidth]{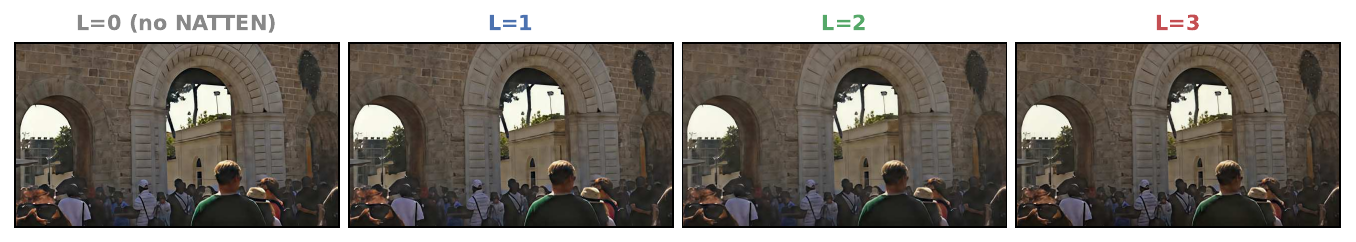}
  \caption{UDM10 clip \texttt{000}: qualitative effect of NATTEN layer count $L$.}
  \label{fig:udm10_000_layer_comparison}
\end{figure}

\begin{figure}[H]
  \centering
  \includegraphics[width=\linewidth]{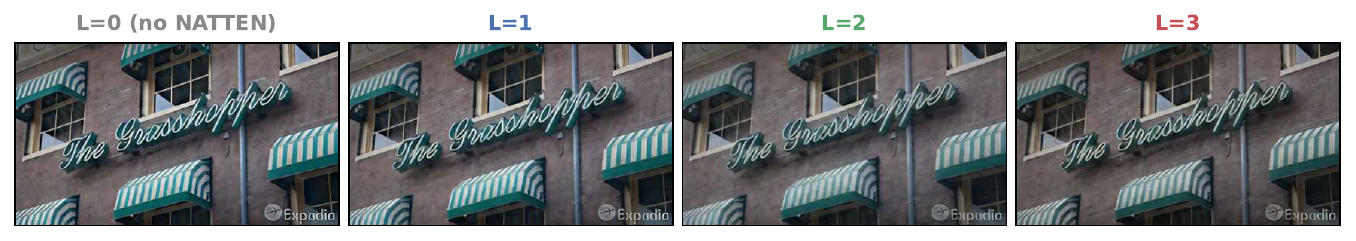}
  \caption{SPMCS clip \texttt{AMVTG\_004}: qualitative effect of NATTEN layer count $L$.}
  \label{fig:spmcs_amvtg004_layer_comparison}
\end{figure}

\begin{figure}[H]
  \centering
  \includegraphics[width=\linewidth]{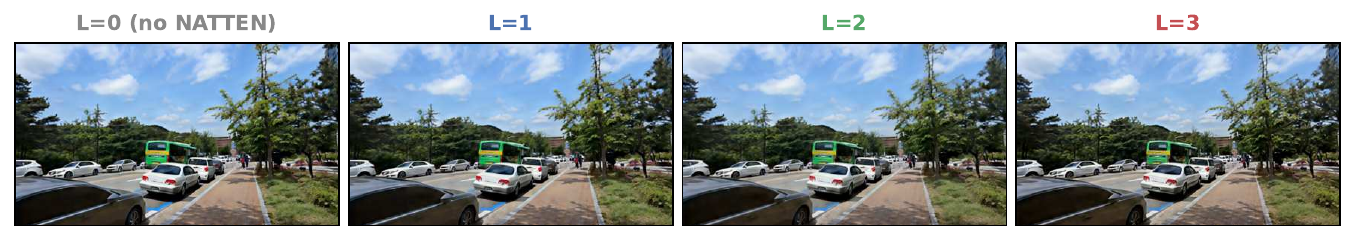}
  \caption{REDS4 clip \texttt{000}: qualitative effect of NATTEN layer count $L$.}
  \label{fig:reds4_000_layer_comparison}
\end{figure}

\begin{figure}[H]
  \centering
  \includegraphics[width=\linewidth]{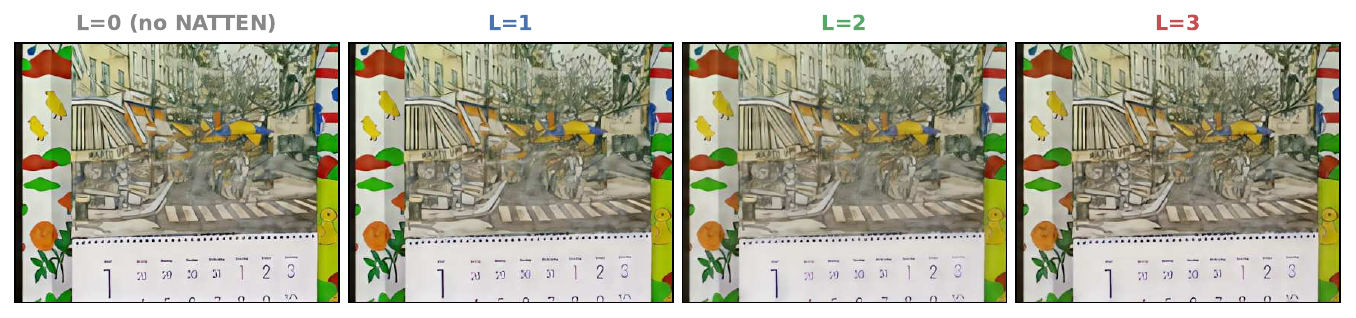}
  \caption{Vid4 clip \texttt{calendar}: qualitative effect of NATTEN layer count $L$.}
  \label{fig:vid4_calendar_layer_comparison}
\end{figure}

\FloatBarrier
\endgroup
\flushbottom
\section{Motion Gate Threshold Selection}
\label{app:motion_gate}

\paragraph{Motion-gate threshold and statistic selection.}
Table~\ref{tab:motion_gate_ablation} first situates the gated schedule against an \emph{ungated} baseline, where every intermediate frame is interpolated regardless of motion magnitude. 
The contrast motivates gating in the first place: ungated inference reaches the highest throughput at both $K{=}2$ and $K{=}3$ (10.83 and 15.16 FPS, respectively) but incurs catastrophic temporal degradation, with tOF reaching 4.088 and 5.226 --- up to 4--6$\times$ worse than the strictest gated setting and far above the $K{=}1$ reference (tOF $=0.818$). 
Some form of motion gating is therefore necessary for high-motion content such as REDS4, rather than merely beneficial. 
Without it, the throughput gained from interpolation comes at a temporal-consistency cost that is an order of magnitude larger than the distortion cost of any threshold choice considered below.

We jointly sweep two candidate motion statistics --- the mean and max Farneback optical-flow magnitude computed on the corresponding low-resolution frames --- each at three thresholds $\tau_{gate}$ chosen to produce comparable fallback tiers (strict/mid/loose) across statistics: $\tau_{gate} \in \{4, 6, 9\}$ for mean-stat and $\tau_{gate} \in \{15, 20, 25\}$ for max-stat.
Within each statistic, quality and temporal-consistency metrics degrade monotonically and consistently as $\tau_{gate}$ increases (i.e., as the tier loosens) for both $K{=}2$ and $K{=}3$: PSNR, SSIM, DISTS, BRISQUE, MUSIQ, MANIQA, $E_{\text{warp}}$, and tOF all worsen as the fallback rate drops. 
The effect is most pronounced in tOF, which for max-stat more than doubles between the strict and loose tiers (1.08 $\to$ 2.53 at $K{=}2$; 0.85 $\to$ 2.78 at $K{=}3$) while PSNR degrades by only 2--4\% over the same range. This asymmetry is consistent with the gate's intended role: it is disproportionately effective at protecting temporal consistency specifically, and this protection erodes quickly once the threshold is relaxed. 
More high-motion frames are interpolated rather than falling back to direct super-resolution. 
We also observe that $K{=}3$ is more sensitive to threshold choice than $K{=}2$ --- SSIM drops by 0.075 across the max-stat sweep at $K{=}3$ versus 0.043 at $K{=}2$, and the fallback rate swings over a wider range (90.9\%$\to$47.0\% vs.\ 74.0\%$\to$38.0\%) --- indicating that larger keyframe intervals rely more heavily on the gate correctly identifying unsafe frames.

Comparing the two statistics at matched tiers isolates the effect of the statistic itself. 
At every tier, mean-stat achieves higher throughput than max-stat (e.g., 6.84 vs.\ 6.62 FPS at the strict tier for $K{=}2$; 9.63 vs.\ 8.17 FPS at the loose tier) and comparable distortion metrics (PSNR, SSIM, $E_{\text{warp}}$), but consistently worse tOF (e.g., 1.302 vs.\ 1.079 at the strict tier; 3.338 vs.\ 2.530 at the loose tier) alongside a lower fallback rate. 
This pattern is consistent with the two statistics capturing different failure modes: a frame can contain a small region of fast motion --- e.g., a moving foreground object against a static background --- that is diluted in the mean but preserved in the peak, so max-stat is more likely to correctly flag such frames as unsafe to interpolate, at the cost of a higher fallback rate and lower throughput. 
Since our design priority is protecting temporal consistency on precisely the high-motion content this gate targets, we adopt max-stat as the default statistic, and within it select $\tau_{gate}=15$: it yields the best values across all quality and temporal-consistency metrics among the max-stat tiers, at the cost of a higher fallback rate than the looser thresholds.
Mean-stat remains a reasonable alternative in deployments where throughput is prioritized over worst-case temporal stability.

\begin{table}[!ht]
  \centering
  \footnotesize
    \caption{Max-stat vs.\ mean-stat motion score compared head-to-head on REDS4, at fixed $L{=}2$, across key-frame intervals $K \in \{2,3\}$.  Fallback \% is the share of frames that exceeded $\tau_{gate}$ and fell back to per-frame SR instead of interpolating. Batch size = 8.}
  \resizebox{\linewidth}{!}{%
  \begin{tabular}{cc c ccccccc ccc cc}
    \toprule
    $K$ & stat & $\tau_{gate}$ & PSNR$\uparrow$ & SSIM$\uparrow$ & LPIPS$\downarrow$ & DISTS$\downarrow$ &
    BRISQUE$\downarrow$ & MUSIQ$\uparrow$ & MANIQA$\uparrow$ &
    $E_{\text{warp}}\downarrow$ & tOF$\downarrow$ & LPIPS$_{\text{inter}}\downarrow$ &
    Fallback\%$\downarrow$ & FPS$\uparrow$ \\
    \midrule
    1 & -- & -- & \textbf{21.59} & \textbf{0.8058} & \textbf{0.3664} & \textbf{0.1074} & \textbf{12.76} & \textbf{69.23} & \textbf{0.4520} & \textbf{0.0349} & \textbf{0.8184} & \textbf{0.3756} & n/a & 5.83 \\
    \midrule
    \multirow{7}{*}{2}
      & -- & ungated & 20.44 & 0.7133 & 0.3904 & 0.1126 & 16.08 & 68.19 & 0.4273 & 0.0510 & 4.0881 & 0.3640 & \textbf{0.0} & \textbf{10.83} \\
      \cmidrule(l){2-15}
      & mean & $\tau_{gate}{=}4$ (strict) & \textbf{21.52} & \textbf{0.7988} & \textbf{0.3681} & 0.1083 & 13.50 & 68.90 & 0.4429 & \textbf{0.0332} & 1.3015 & 0.3695 & 68.0 & 6.84 \\
      & max  & $\tau_{gate}{=}15$ (strict) & 21.50 & 0.7960 & 0.3683 & \textbf{0.1082} & \textbf{13.41} & \textbf{68.99} & \textbf{0.4449} & 0.0341 & \textbf{1.0788} & 0.3711 & 74.0 & 6.62 \\
      \cmidrule(l){2-15}
      & mean & $\tau_{gate}{=}6$ (mid) & 21.21 & 0.7730 & 0.3744 & 0.1097 & 14.25 & 68.65 & 0.4373 & 0.0380 & 2.1588 & 0.3648 & 45.0 & 7.81 \\
      & max  & $\tau_{gate}{=}20$ (mid) & 21.23 & 0.7746 & 0.3736 & 0.1094 & 14.06 & 68.81 & 0.4406 & 0.0382 & 1.7905 & 0.3680 & 56.0 & 7.31 \\
      \cmidrule(l){2-15}
      & mean & $\tau_{gate}{=}9$ (loose) & 20.73 & 0.7349 & 0.3843 & 0.1115 & 15.46 & 68.36 & 0.4307 & 0.0458 & 3.3382 & \textbf{0.3631} & 14.5 & 9.63 \\
      & max  & $\tau_{gate}{=}25$ (loose) & 20.96 & 0.7527 & 0.3793 & 0.1104 & 14.71 & 68.63 & 0.4367 & 0.0428 & 2.5302 & 0.3653 & 38.0 & 8.17 \\
    \midrule
    \multirow{7}{*}{3}
      & -- & ungated & 19.57 & 0.6490 & 0.4097 & 0.1160 & 15.92 & 67.92 & 0.4183 & 0.0556 & 5.2260 & \textbf{0.3455} & \textbf{0.0} & \textbf{15.16} \\
      \cmidrule(l){2-15}
      & mean & $\tau_{gate}{=}4$ (strict) & 21.47 & 0.7975 & 0.3687 & 0.1086 & 13.28 & 68.91 & 0.4434 & \textbf{0.0336} & 1.4048 & 0.3655 & 75.8 & 6.85 \\
      & max  & $\tau_{gate}{=}15$ (strict) & \textbf{21.57} & \textbf{0.8039} & \textbf{0.3666} & \textbf{0.1076} & \textbf{12.95} & \textbf{69.12} & \textbf{0.4489} & 0.0339 & \textbf{0.8481} & 0.3724 & 90.9 & 6.17 \\
      \cmidrule(l){2-15}
      & mean & $\tau_{gate}{=}6$ (mid) & 21.31 & 0.7850 & 0.3720 & 0.1094 & 13.56 & 68.79 & 0.4396 & 0.0346 & 1.7312 & 0.3602 & 63.6 & 7.51 \\
      & max  & $\tau_{gate}{=}20$ (mid) & 21.40 & 0.7895 & 0.3699 & 0.1084 & 13.26 & 68.96 & 0.4440 & 0.0348 & 1.1600 & 0.3660 & 75.8 & 6.85 \\
      \cmidrule(l){2-15}
      & mean & $\tau_{gate}{=}9$ (loose) & 20.50 & 0.7221 & 0.3888 & 0.1124 & 14.75 & 68.36 & 0.4287 & 0.0440 & 3.3532 & 0.3511 & 31.1 & 10.12 \\
      & max  & $\tau_{gate}{=}25$ (loose) & 20.63 & 0.7289 & 0.3867 & 0.1115 & 14.26 & 68.61 & 0.4350 & 0.0437 & 2.7835 & 0.3557 & 47.0 & 8.65 \\
    \bottomrule
    \end{tabular}}
    \label{tab:motion_gate_ablation}
\end{table}

\FloatBarrier
\section{SOTA Comparison}
\label{app:sota}
\raggedbottom
\begingroup
\setlength{\intextsep}{6pt plus 2pt minus 2pt}
\setlength{\abovecaptionskip}{-6pt}
\setlength{\belowcaptionskip}{0pt}

Uncurated zoomed in comparisons across all datasets and approaches.

\begin{figure}[H]
  \centering
  \includegraphics[width=\linewidth]{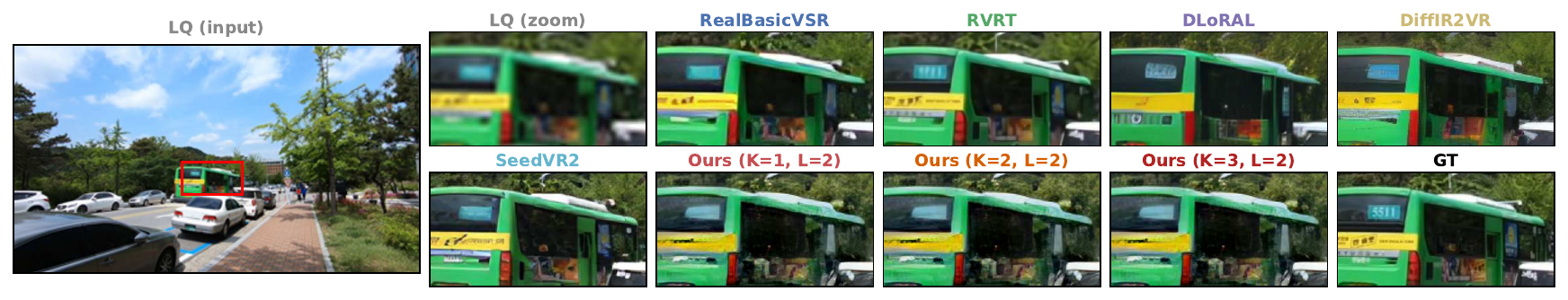}
  \caption{REDS4 clip \texttt{000}: frame-level comparison against baselines.}
  \label{fig:reds4_000_comparison}
\end{figure}

\begin{figure}[H]
  \centering
  \includegraphics[width=\linewidth]{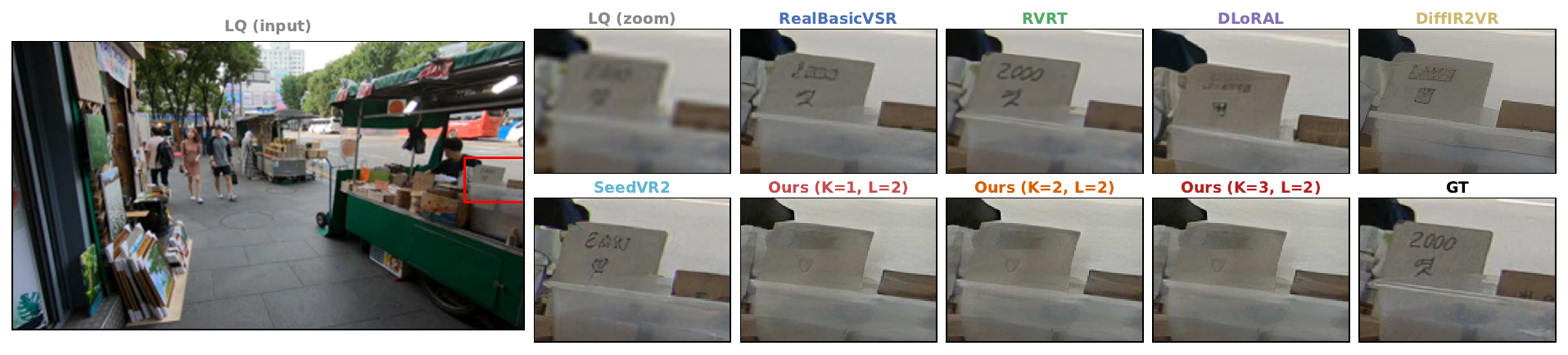}
  \caption{REDS4 clip \texttt{020}: frame-level comparison against baselines.}
  \label{fig:reds4_020_comparison}
\end{figure}

\begin{figure}[H]
  \centering
  \includegraphics[width=\linewidth]{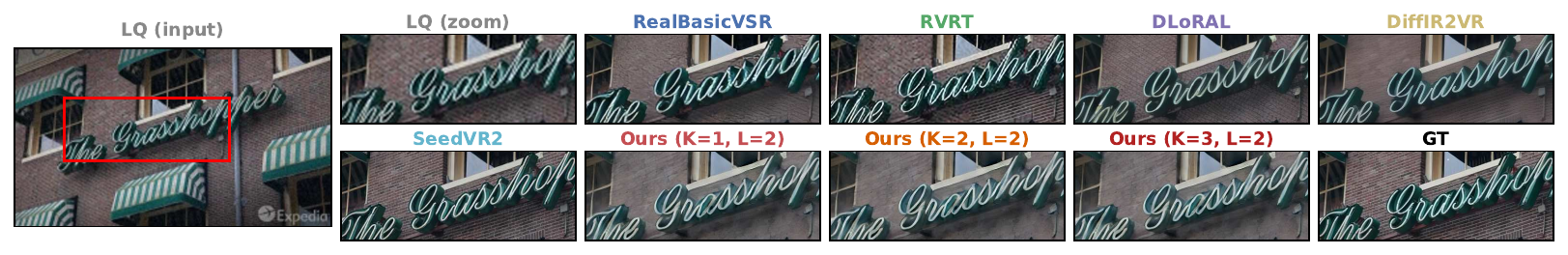}
  \caption{SPMCS clip \texttt{AMVTG\_004}: frame-level comparison against baselines.}
  \label{fig:spmcs_amvtg004_comparison}
\end{figure}

\begin{figure}[H]
  \centering
  \includegraphics[width=\linewidth]{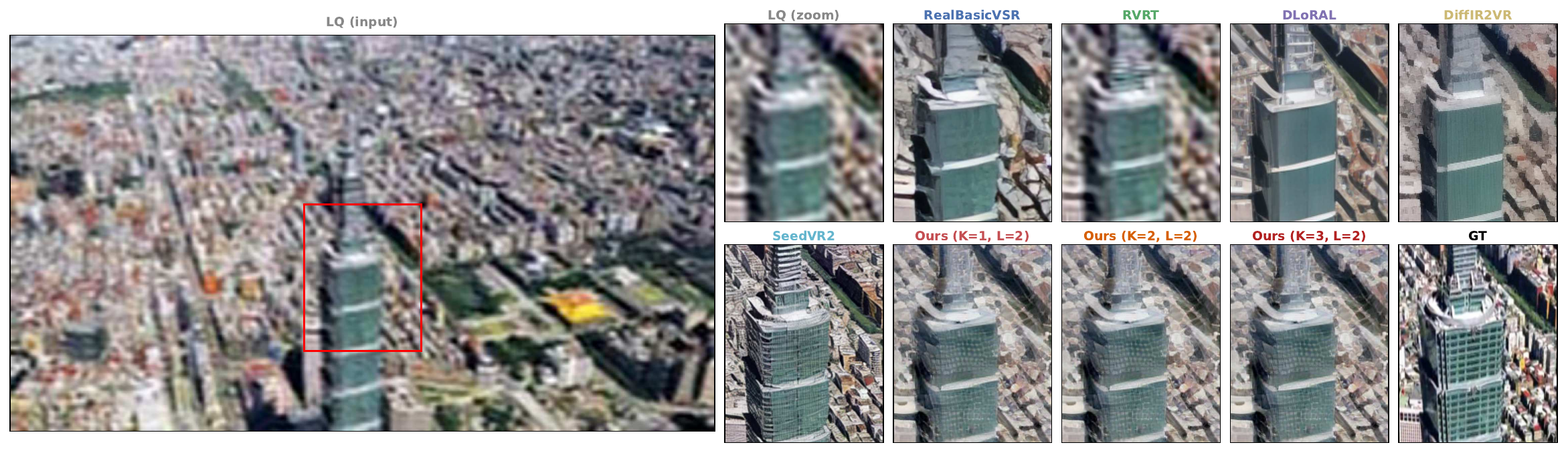}
  \caption{SPMCS clip \texttt{hdclub\_003\_001}: frame-level comparison against baselines.}
  \label{fig:spmcs_hdclub003001_comparison}
\end{figure}

\begin{figure}[H]
  \centering
  \includegraphics[width=\linewidth]{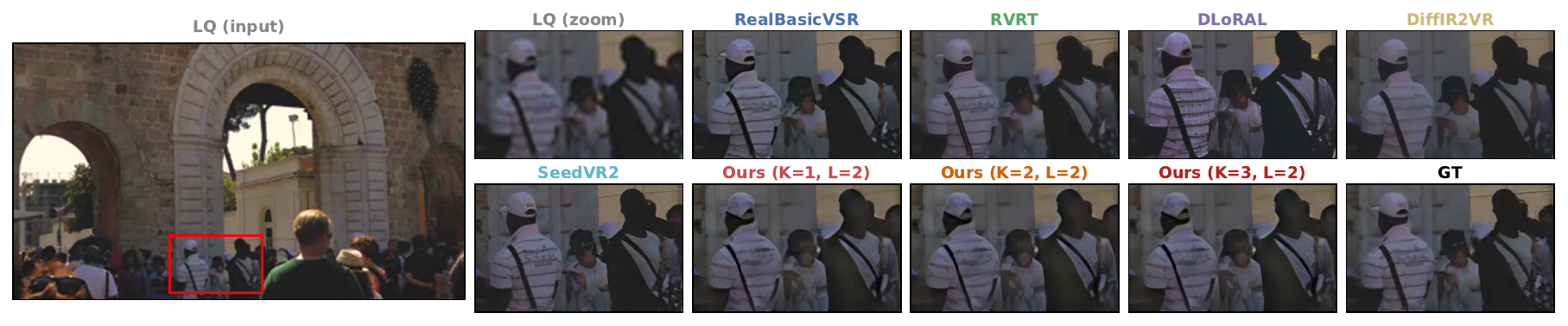}
  \caption{UDM10 clip \texttt{000}: frame-level comparison against baselines.}
  \label{fig:udm10_000_comparison}
\end{figure}

\begin{figure}[H]
  \centering
  \includegraphics[width=\linewidth]{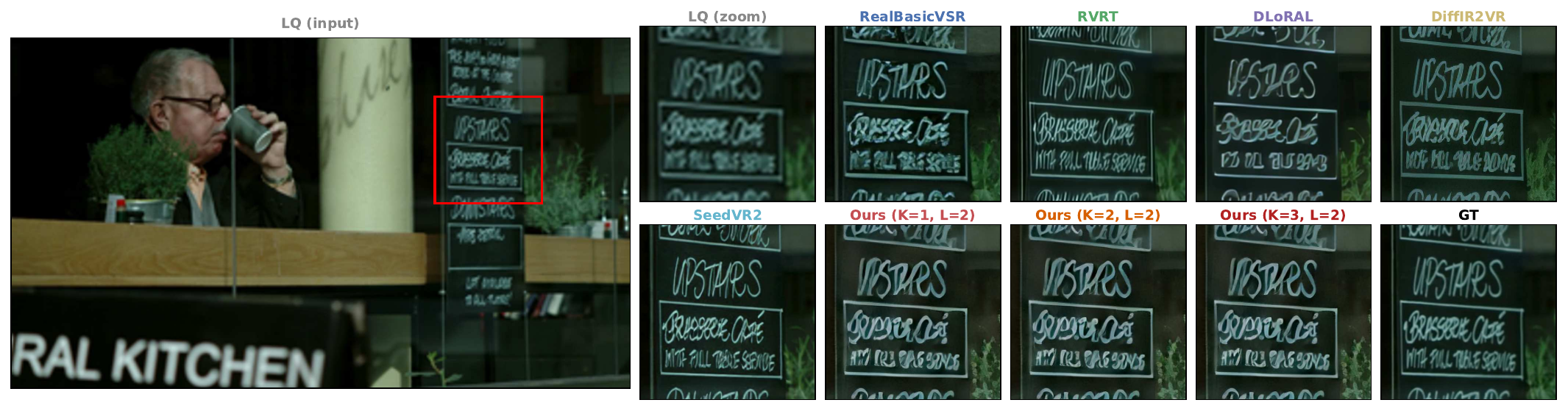}
  \caption{UDM10 clip \texttt{004}: frame-level comparison against baselines.}
  \label{fig:udm10_004_comparison}
\end{figure}

\begin{figure}[H]
  \centering
  \includegraphics[width=\linewidth]{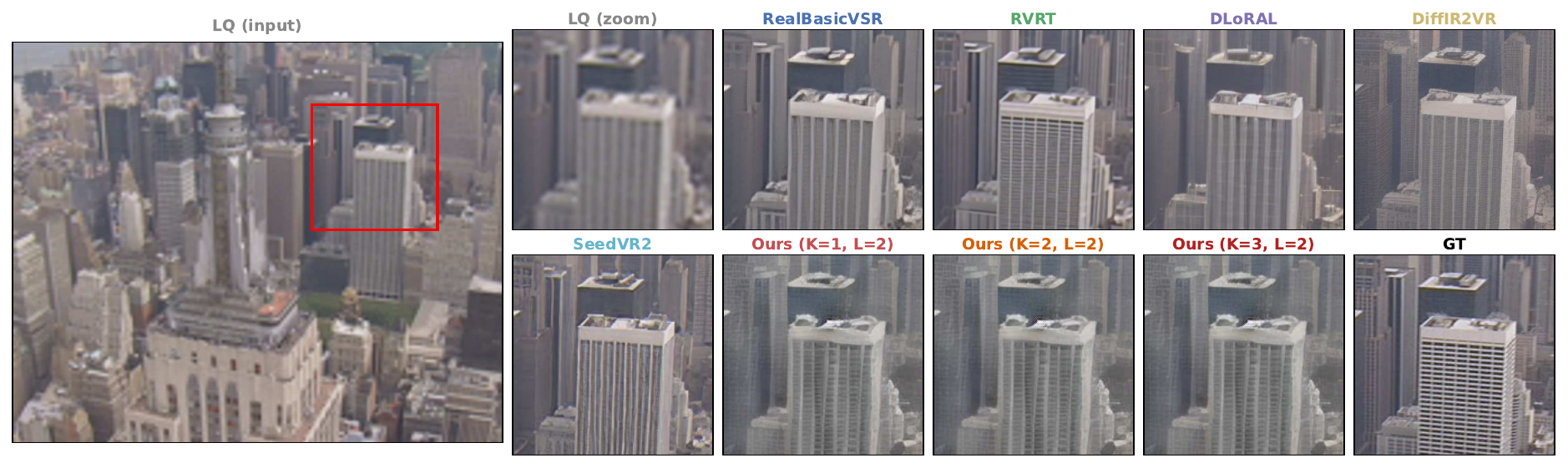}
  \caption{Vid4 clip \texttt{city}: frame-level comparison against baselines.}
  \label{fig:vid4_city_comparison}
\end{figure}

\begin{figure}[H]
  \centering
  \includegraphics[width=\linewidth]{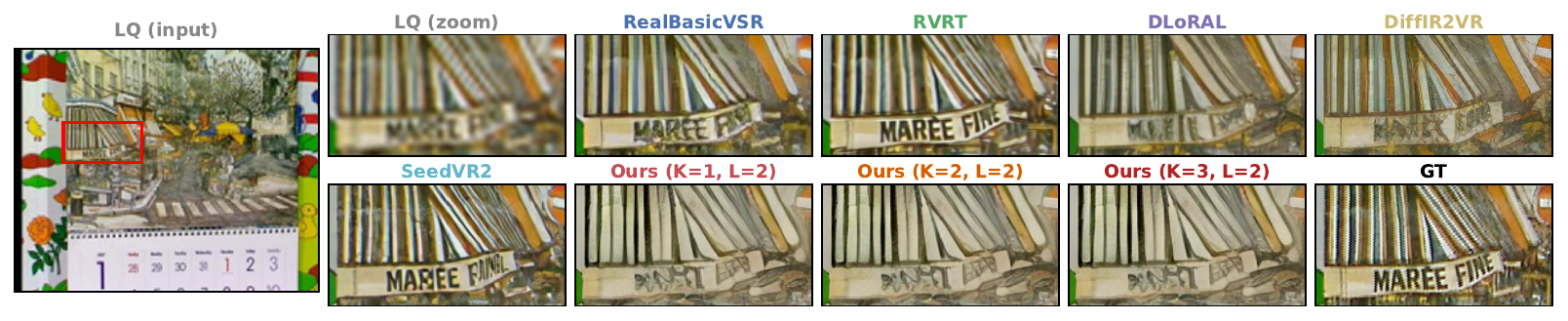}
  \caption{Vid4 clip \texttt{calendar}: frame-level comparison against baselines.}
  \label{fig:vid4_calendar_comparison}
\end{figure}

\FloatBarrier
\endgroup
%

\FloatBarrier
\end{document}